\documentclass[pdflatex,sn-mathphys-num]{sn-jnl}% Math and Physical Sciences Numbered Reference Style 
\usepackage{tabularray}%
\usepackage{rotating}

\usepackage{graphicx}%
\usepackage{multirow}%
\usepackage{amsmath,amssymb,amsfonts}%
\usepackage{amsthm}%
\usepackage{mathrsfs}%
\usepackage[title]{appendix}%
\usepackage{textcomp}%
\usepackage{manyfoot}%
\usepackage{booktabs}%
\usepackage{algorithmicx}%
\usepackage{algpseudocode}%
\usepackage{listings}%

\usepackage{multicol}
\usepackage[table,xcdraw]{xcolor}
\usepackage[ruled,vlined,linesnumbered]{algorithm2e}
\usepackage{lmodern}
\usepackage{setspace}
\usepackage{caption}
\usepackage{subcaption}
\SetEndCharOfAlgoLine{}
\SetKwComment{Comment}{/* }{ */}

\definecolor{sky}{RGB}{202, 240, 248}
\theoremstyle{thmstyleone}%
\theoremstyle{thmstyletwo}%

\theoremstyle{thmstylethree}%

\begin{document}

\title[Time-series Meets Complex Motion Modeling:\\Robust and Computational-effective Motion Predictor for Multi-object Tracking]{Time-series Meets Complex Motion Modeling:\\Robust and Computational-effective Motion Predictor for Multi-object Tracking}

\author[1,2]{\fnm{Nhat-Tan} \sur{Do}}\email{21522575@gm.uit.edu.vn}
\author[1,2]{\fnm{Le-Huy} \sur{Tu}}\email{21522173@gm.uit.edu.vn}
\author[1,2]{\fnm{Nhi Ngoc-Yen} \sur{Nguyen}}\email{21521231@gm.uit.edu.vn}
\author[1,2]{\fnm{Dieu-Phuong} \sur{Nguyen}}\email{21520091@gm.uit.edu.vn}
\author*[1,2]{\fnm{Trong-Hop} \sur{Do}}\email{hopdt@uit.edu.vn}
\affil*[1]{\orgdiv{Faculty of Information Science and Engineering}, \orgname{University of
 Information Technology}, \orgaddress{\city{Ho Chi Minh City}, \country{Vietnam}}}
\affil[2]{\orgname{Vietnam National University}, \orgaddress{\city{Ho Chi Minh City}, \country{Vietnam}}}

\abstract{Multi-object tracking (MOT) is critical in numerous real-world applications, including surveillance, autonomous driving, and robotics. Accurately predicting object motion is fundamental to MOT, but current methods struggle with the complexities of real-world, non-linear motion (e.g., sudden stops, sharp turns).
While recent research has gravitated towards increasingly complex and computationally expensive generative models to tackle this problem, their practical utility is often constrained. This paper challenges that paradigm, arguing that such complexity is not only unnecessary but can be outperformed by a more efficient, purpose-built approach.
We introduce the \textbf{T}emporal \textbf{C}onvolutional \textbf{M}otion \textbf{P}redictor (TCMP), a novel framework for MOT that leverages a modified Temporal Convolutional Network (TCN) featuring dilated convolutions and a regression head. This design allows for effective motion prediction across arbitrary temporal context lengths.
Experimental results demonstrate that our approach achieves state-of-the-art performance, specifically improves upon the previous best method in several key metrics: HOTA (a measure of overall tracking accuracy) increases from 62.3\% to 63.4\%, IDF1 (a measure of identity preservation) rises from 63.0\% to 65.0\%, and AssA (a measure of association accuracy) improves from 47.2\% to 49.1\%.
Significantly, TCMP achieves this performance while being highly efficient; it has only 0.014 times the parameters and requires only 0.05 times the computational cost (FLOPs) compared to the SOTA method.
while is only 0.014 times the size (in terms of parameters) and requires only 0.05 times the computational cost (in terms of FLOPs). These findings highlight the robustness of our method to advance MOT systems by ensuring adaptability, accuracy, and efficiency in complex tracking environments.}




\maketitle

\section{Introduction}
\label{sec:intro}

\begin{figure*}[t]
    \centering
    \includegraphics[width=1\linewidth]{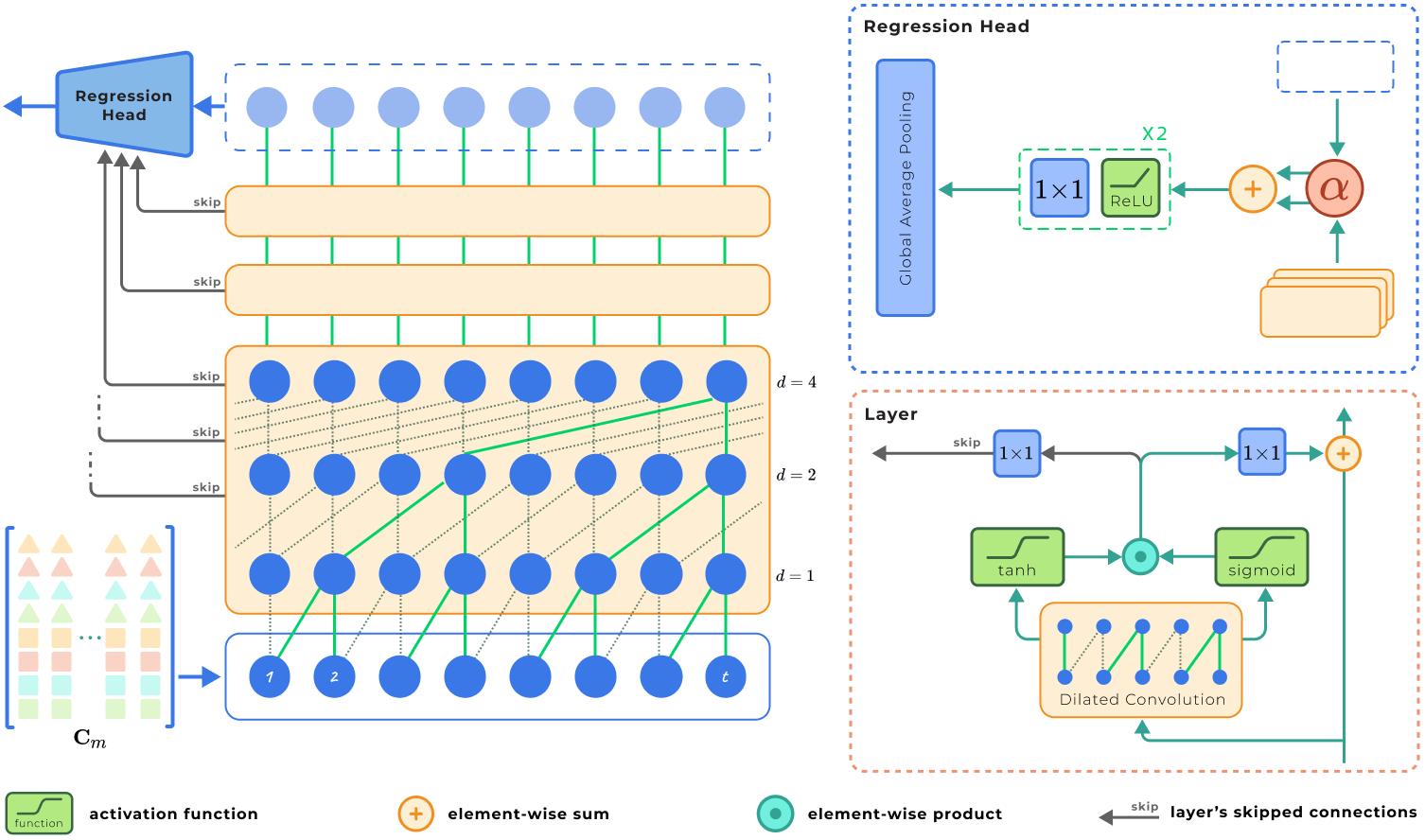}
    \caption{The overall architecture of our proposed Temporal Convolutional Motion Predictor}
    \label{fig:model-architecture}
\end{figure*}

Multi-object Tracking (MOT) is a critical task in computer vision that involves detecting and tracking multiple objects across frames in a video sequence. The task has many real-life applications, including autonomous driving, surveillance, sports analytics, and robotics. The ability to track multiple objects accurately in real-time is essential for these applications to function effectively and safely.

Multi-object Tracking can be broadly categorized into linear and non-linear motion. Linear motion, where objects move at relatively constant speeds and directions (e.g., pedestrians walking on a straight sidewalk), is considerably easier to model and predict. In these scenarios, the Kalman Filter, a well-known classical approach in MOT, has been proven to work effectively even under the assumption of linear motion. However, real-world scenarios frequently involve non-linear motion. Examples include objects suddenly stopping or changing speed, sharp turns, complex interactions, or occlusions. Modeling these unpredictable movements significantly increases the complexity of the tracking problem. Addressing this non-linearity is a major focus of ongoing research in MOT.

To advance research on non-linear motion prediction, specialized benchmarks like DanceTrack \cite{sun2022dancetrack}, and SportsMOT \cite{cui2023sportsmot} have been introduced. These datasets present complex and dynamic environments and involve tracking objects with unpredictable and non-linear trajectories, such as dancers or athletes. Unlike the relatively predictable motion found in pedestrian-focused datasets \cite{milan2016mot16, dendorfer2020mot20}, DanceTrack and SportsMOT pose significant challenges due to rapid changes in direction, speed, and complex interactions.
Another challenge in DanceTrack and SportsMOT is that the objects often wear uniforms or have similar appearances, making the Re-ID module perform poorly. The varying difficulty levels in these benchmarks challenge the robustness and adaptability of tracking algorithms. In such cases, the Kalman Filter is less effective, as it struggles to handle the non-linear motion patterns.

Despite significant advancements, one of the critical problems of motion prediction in Multi-Object Tracking (MOT) lies in balancing model complexity and performance. Recent advancements often achieve superior performance with complex and large models. For instance, \citet{lv2024diffmot} proposes a decoupled diffusion-based motion predictor (D$^2$MP), a diffusion-based motion prediction model that achieves impressive results on non-linear benchmarks like DanceTrack \cite{sun2022dancetrack} and SportsMOT \cite{cui2023sportsmot}. While achieving notable results, the design complexity and computational demands of D$^2$MP are substantial. On the other hand, many efforts, both utilizing \cite{chaabane2021deft, xiao2024motiontrack, zhang2020multiple} or not utilizing a neural network \cite{caoOCSort2023}, although not as complicated, are not as successful when compared to larger models.

This work proposes a novel \textbf{T}emporal \textbf{C}onvolutional \textbf{M}otion \textbf{P}redictor (TCMP) that effectively addresses this gap. By leveraging the efficiency of convolutional operations in processing arbitrary-range temporal information, our method achieves superior performance while maintaining an exceptionally lightweight architecture.
Specifically, TCMP serves as an advanced motion prediction module designed to integrate into standard tracking-by-detection frameworks (e.g., those using SORT or BYTE association). By generating more accurate motion estimates, particularly for non-linear trajectories, TCMP provides higher quality inputs to the crucial data association stage, thereby improving overall tracking performance.
Theoretically, TCMP offers superior performance by directly addressing the causality of motion prediction in online MOT. It focuses simply on one-step-ahead forecasting. This approach avoids the cumulative errors and computational inefficiencies often associated with autoregressive modeling.
TCMP leverages dilated convolutions within a Temporal Convolutional Network (TCN) to efficiently capture temporal dependencies in historical data, expanding the context length the model can capture without expensive computational costs.
We demonstrate that TCMP not only surpasses prior state-of-the-art models in key benchmarks such as DanceTrack, MOT, and SportsMOT but also does so with a fraction of the computational complexity and parameter count. These findings underscore the potential of smaller, simpler models to redefine the state of the art in MOT.

 Our contribution can be summarized as follows:
 \begin{itemize}
     \item We provide an in-depth analysis of the causality in motion prediction for online MOT, justifying our decision to focus on one-step-ahead forecasting rather than autoregressive modeling. This approach avoids cumulative errors and computational inefficiencies, offering a more robust framework for motion prediction.
     
     \item We propose a novel \textbf{T}emporal \textbf{C}onvolutional \textbf{M}otion \textbf{P}redictor (TCMP) for motion prediction in MOT, designed to be lightweight, balance simplicity and performance. To the best of our knowledge, our work is the first to fully use a temporal convolutional network as a motion prediction model. By employing convolutional operations for temporal modeling, TCMP achieves SOTA performance with significantly fewer parameters and lower FLOPs compared to previous SOTA, demonstrating that smaller models can outperform more complex architectures.

     \item Through comprehensive experiments on diverse, challenging benchmarks (DanceTrack and SportsMOT), we demonstrate our model's superior performance in handling challenging scenarios, including occlusions, missed detections, and complex motions. Qualitative comparisons highlight its ability to maintain object identities under severe occlusion and trajectory crossing.
 \end{itemize}

\section{Related Works}
\label{sec:related_works}

Multi-object tracking (MOT) has advanced significantly through progress in object detection,  motion modeling, and data association. Motion models predict the future positions of objects, a core component of MOT. A robust motion prediction model helps significantly improve the overall MOT performance since it can (1) reduce the impact of noisy detection and (2) produce a reliable input for the association step. Motion models fall into two broad categories: linear and non-linear. While linear motion models like the Kalman Filter are efficient for simple trajectories, real-world scenarios frequently involve non-linear motion (e.g., an object suddenly stopping or changing speed, sharp turns, complex interactions, or occlusions), significantly increasing the tracking problem's complexity.

\subsection{MOT in Linear Motion Scenarios}

The rapid development of object detection \cite{cai2018cascade, sun2021makes, ge2021yolox} has contributed significantly and made Tracking-by-detection become the dominant paradigm in the problem of Multi-object Tracking in recent years. In the Tracking-by-detection paradigm, most current methodologies are built upon the simple SORT framework, which uses Kalman Filter (KF) to perform state estimation in $2$-dimensional image space and frame-by-frame data association using the IoU (Intersect-over-Union) metric with the Hungarian algorithm. KF works as a Motion prediction model in the SORT framework.  Algorithms like SORT \cite{bewleySORT2016}, DeepSORT \cite{wojkeDeepSORT2017}, FairMOT \cite{zhangFairMOT2021}, ByteTrack \cite{zhangByteTrack2022}, and OC-SORT \cite{caoOCSort2023} utilize KF, assuming constant object velocity and direction over short intervals. 

\citet{zhangByteTrack2022} point out that the performance bottleneck is the detection part, not the association part. Therefore, a robust detector with a simple association strategy still leads to good tracking results. Many efforts \cite{zhou2019objects, lu2020retinatrack, wang2020towards, peng2020chained, zhangFairMOT2021} try to fuse detection and tracking into a unified framework. 

After the detection step, accurately match tracks from previous frames with detection boxes in the current frame. This step, in the tracking-by-detection paradigm, is the association step. SORT \cite{bewleySORT2016} pioneered a lightweight association method using motion cues. While efficient, its simplicity can falter under challenging conditions like occlusions or camera motion. Extensions such as DeepSORT \cite{wojkeDeepSORT2017} incorporated appearance cues, enhancing robustness in complex environments. Similarly, MOTDT \cite{chenRealtimeMultiplePeople2018} leveraged appearance-based matching before relying on motion cues for residual associations. 
Further advancements like BoTSORT \cite{aharonBoTSORT2022} refined these strategies with camera motion compensation and improved distance metric fusion, achieving state-of-the-art results in scenarios involving occlusions and missed detections. These innovations collectively enhance MOT's reliability, addressing key challenges in real-world applications.

FairMOT \cite{zhangFairMOT2021} unified detection and re-identification (ReID) tasks within a single network, addressing optimization imbalances and improving feature representation. This approach proved effective in dense scenes. ByteTrack \cite{zhangByteTrack2022} introduced a confidence-based association strategy, utilizing both high- and low-confidence detections to handle occlusions and motion blur.

In linear motion scenarios, these methods, particularly those incorporating appearance information and robust association techniques, have proven highly effective, achieving strong performance on benchmarks featuring predictable object trajectories.

\subsection{MOT in Non-linear Motion Scenarios}

While the Kalman Filter and related methods excel in linear motion scenarios, they struggle to accurately model the complexities of \textit{non-linear} motion, which is prevalent in real-world environments. Non-linear motion involves abrupt changes in speed and direction, complex interactions between objects, and frequent occlusions, making prediction significantly more difficult.

To address this challenge, more advanced motion models have been developed. Optical flow-based models \cite{chenRealtimeMultiplePeople2018} calculate pixel displacements between frames to derive motion information. RNN-based and LSTM-based models \cite{milan2017online, sadeghian2017tracking, wan2018online, ran2019robust, chaabane2021deft} leverage sequence modeling in latent space, capturing temporal dependencies. Transformer-based models \cite{sunTransTrack2020, zengMOTR2022, cai2024iouformer} utilize attention mechanisms for long-range motion dependencies, offering richer representations. These methods require a large amount of training data, expensive computational resources, and, as a result, long training times. 

Despite these advancements, accurately and efficiently predicting non-linear motion in MOT remains a significant open challenge, and is the primary focus of our work.

\section{Foundation of Motion Prediction}
\label{sec:methodology}

\subsection{Problem Statement}
\label{sec:problem-statement}

We are interested in tracking objects by locating their bounding boxes. Formally, we define $\mathcal{O}_t$ the object $\mathcal{O}$ at time $t$, the bounding box $\mathbf{B_t} = (x_t, y_t, w_t, h_t)^\top \in \mathbb{R}^{4}$ represents the coordinate in $2$-dimensional space, width and height of $\mathcal{O}_t$, respectively. Let $\mathbf{T}_{M} = ( \mathbf{B}_1, \dots,  \mathbf{B}_t, \dots, \mathbf{B}_m ) \in \mathbb{R}^{4 \times m}$ represents the trajectory of object $\mathcal{O}$ from time $1 \le t \le m$. Note that $m$ may vary for each object $\mathcal{O}$, since all objects may not appear simultaneously. We define $\mathbf{M}_t \in \mathbb{R}^4$ as the "motion" of the object $\mathcal{O}$ in a given time $t$. This motion presentation is calculated as the difference between the bounding box in the frame at time $t$ and the previous frame $t-1$:
\begin{equation}
\label{eq:e_m_ot}
\mathbf{M}_t = \mathbf{B}_t - \mathbf{B}_{t-1} = (\Delta x_t, \Delta y_t, \Delta w_t, \Delta h_t)
\end{equation}

As we stack $\mathbf{B}_t$ with its corresponding $\mathbf{M}_t$ together, we form the Context $\mathbf{C}_t \in \mathbb{R}^{8}$ as following:
\begin{equation}
\label{eq:c_t}
\mathbf{C}_t = (x_t,y_t, w_t, h_t, \Delta x_t, \Delta y_t, \Delta w_t, \Delta h_t)^\top
\end{equation}

Thus, $\mathbf{C}_{m} = \left[ \mathbf{C}_1, \dots, \mathbf{C}_t, \dots , \mathbf{C}_m \right] \in \mathbb{R}^{8 \times m}$ represents the temporal information of $\mathcal{O}$ up to $m$.
It is necessary to define $\mathbf{C}_{m}$ as it is proven in \citet{lv2024diffmot} that (1) using the positional information from bounding box $\mathbf{B}_{t-1}$ only can lead to significant deviations of prediction, and (2) using the motion change information $\mathbf{M}_{t-1}$ only can introduce difficulty in generation.

Using this defined contextual representation, the motion prediction problem can be framed as modeling a function $f: \mathbb{R}^{8 \times m} \rightarrow \mathbb{R}^{4}$ parameterized by $W$ that map the given context $\mathbf{C}_{m}$ to a prediction $\widehat{\mathbf{M}}_{m+1}$:
\begin{equation}
    \widehat{\mathbf{M}}_{m+1} =  f(\mathbf{C}_{m}, W) 
\end{equation}

This formula aligns with the causality of Online Multi-object Tracking scenarios, where the estimation in time $t = m+1$ only depends on information from previous time steps $t \in \{1, 2, \dots, m\}$.

We empirically design the parameters $W$ as a set of dilated convolution kernels within a Temporal Convolutional Network (TCN), a Regression Head, and a parameter $\alpha$. The following section describes the detailed architecture and shows the effectiveness of this parameterized approach in multi-object tracking. 

\subsection{Analysis of the Causality of Motion Prediction in Multi-Object Tracking (MOT)}

Motion prediction in MOT is a causal framework: The prediction at any time step $t+1$ depends solely on information available up to time $t$, and future observations do not influence current prediction. 
This causality introduces several considerations:
\begin{itemize}
    \item \textbf{Processing information causally}. The direct consideration is about the capability of the model to accurately predict the next motion without any access to future observations. As we state the problem in Section \ref{sec:problem-statement}, our model is causal by predicting $\widehat{\mathbf{M}}_{m+1}$ using only $\mathbf{C}_{m}$.

    \item \textbf{Representing Temporal Information}. A critical challenge in causal motion prediction is effectively capturing the temporal dependencies in historical data. Models must leverage past observations to infer motion patterns while respecting the causal constraints of the problem. We solve this problem by leveraging a stack of dilated convolution kernels, expanding the context length the model can capture without expensive computational costs. Further explanation is in Section \ref{subsec:dilated-conv}.

    \item \textbf{Error Accumulation}. In causal systems, prediction errors at time $t+1$ can add up to subsequent predictions, potentially leading to drift over time. The model design needs a mechanism to mitigate error accumulation and maintain robustness over time. We solve this problem by modeling TCMP as a regression model instead of an autoregressive model. Further explanation is in Section \ref{sec:temporal-conv-as-motion-predictor}.

\end{itemize}

\section{Proposed Temporal Convolutional Motion Predictor}

In this section, we describe our proposed method of using a Temporal Convolutional Network as a motion predictor, with extra modifications to make it suitable for the problem of motion prediction for Multi-object Tracking. 

\subsection{Overall Architecture of Temporal Convolutional Motion Predictor}

The overall architecture of our proposed Temporal Convolutional Motion Predictor (TCMP) is illustrated in Figure \ref{fig:model-architecture}. Our model is built upon a Temporal Convolutional Network (TCN). This TCN serves as the core component for modeling temporal dependencies in motion prediction. The TCN effectively captures long-range temporal patterns and non-linear dynamics across object trajectories, providing a rich and robust representation of motion features. A lightweight Regression Head subsequently processes these extracted features to accurately predict the future motion $\mathbf{M}_{m+1}$. This seamless integration of the TCN ensures that the motion prediction leverages both the strength of temporal modeling and computational efficiency.

\subsection{Temporal Convolutional Network for motion predictor}
\label{sec:temporal-conv-as-motion-predictor}

\subsubsection{Dilated Convolution and Traditional Temporal Convolutional Network} \label{subsec:dilated-conv}

\begin{figure*}[t!]
     \centering

    \begin{subfigure}[b]{0.82\textwidth}
        \centering
        \includegraphics[width=\textwidth]{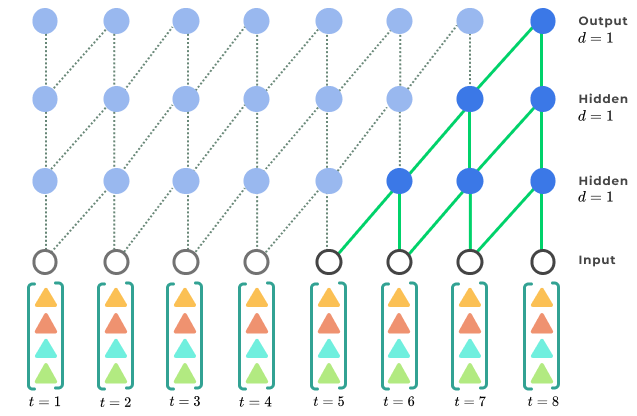}
        \caption{stack of \textit{standard} causal convolutional layers}
        \label{fig:stack-of-normal-causal}
    \end{subfigure} 
    \hfill
    \begin{subfigure}[b]{0.82\textwidth}
        \centering
        \includegraphics[width=\textwidth]{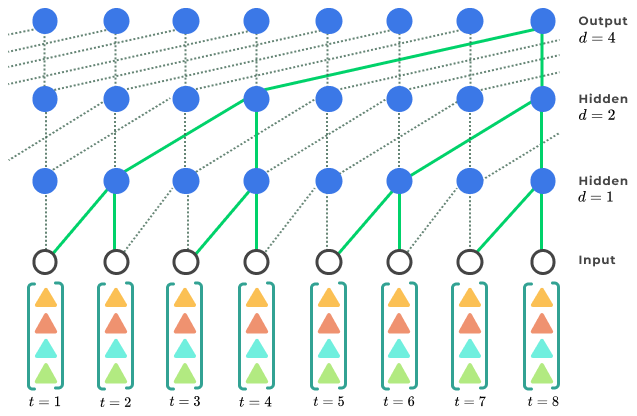}
        \caption{stack of \textit{dilated} causal convolutional layers}
        \label{fig:stack-of-dilated-causal}
    \end{subfigure}
        
    \caption{Visualization of a stack of \textit{standard} causal convolutional layers and a stack of \textit{dilated} causal convolutional layers. Lines indicate the temporal dependencies contributing to the receptive field for each layer, showing which input steps influence later computations. With the same $3$ layers of convolution kernels, the \textit{standard} stack only captures a receptive field of size of $4$, while the \textit{dilated} counterpart captures a context length of up to $8$ with only 3 layers.}
        \label{fig:compare-causal}
\end{figure*}

A standard convolution of 2 functions $x$ and $w$ results in a new function $s$ as follows:

\begin{equation}
    \label{eq:standard-conv}
    s(a) = (x \ast w)(a) = \sum_{t}{x(a-t) \cdot w(t)}
\end{equation}

In the context of motion prediction, Equation \ref{eq:standard-conv} can be explained as follows:
$s(a)$ is the output at a specific time step $a$ after applying the convolution operation. $s(a)$ can be thought of as \textit{the predicted motion or position} at time step $a$. $x$ is the input signal or time series. In this paper, $x$ corresponds to the contextual information $\mathbf{C}$, which includes the bounding box coordinates and motion information of the object. Thus, $x(a - t)$ represents the context at a $t$ time before $a$. $w$ represents the convolution kernel or filter. Here, $w$ represents the parameterized kernels $W$ within the TCN. $t$ is the index variable for the summation, representing the span of the kernel. $x \ast w$ is the convolution operation between the input signal and the kernel.

Dilated convolution is an expanded version of the convolution operation in Equation \ref{eq:standard-conv} that inserts gaps between the kernel elements. Formally, A dilated convolution with the dilation factor $d$ can be defined as follows:
\begin{equation}
    \label{eq:dilated-conv}
    s(a) = (x \ast_d w)(a) = \sum_t x(a- d \cdot t) \cdot w(t)
\end{equation}

This simple formula plays a critical role in our design choice. Systematically, we can expand the receptive field of our model by stacking multiple dilated convolution kernels together. For instance, as shown in Figure \ref{fig:stack-of-dilated-causal}, a stack of $3$ $2 \times 2$ dilated convolution kernels with dilated factor $1, 2, 4$, respectively, can capture a receptive field of size $8$. Consequently, in the context of motion prediction, with the same stack, our model has a context length of $m=8$. For comparison, its counterpart - a stack of standard convolution kernels (Figure \ref{fig:stack-of-normal-causal}) - with the same number of layers can capture only $m=4$ context information.

Temporal Convolutional Network is a stack of dilated convolution layers. The nature of "stack" and "dilated convolution" plays a crucial role in our design choice for building a computationally efficient yet robust motion prediction model. As stated in Section \ref{subsec:dilated-conv}, a systematic stack of dilated convolution kernels can expand the receptive field with just a few layers, eliminating the need to increase complexity or computational cost. Theoretically, the dilation factor increases exponentially in depth, resulting in an exponentially large receptive field. A block of kernels with dilation factors of $1, 2, 4$ has a receptive field of size $8$ and is more efficient and non-linear compared to the counterpart $1 \times 8$ standard convolution.

\subsubsection{Problem of traditional Temporal Convolutional Network}

Theoretically, we can model a Temporal Convolutional Network (TCN) as a generator or an autoregressive model that predicts the box $\mathbf{B}_{m+1}$ using previous boxes $( \mathbf{B}_1, \dots,  \mathbf{B}_t, \dots, \mathbf{B}_m)$, then continue predicting $\mathbf{B}_{m+2}$ in the autoregressive fashion. There are two reasons that we choose not to use TCN this way.
\begin{itemize}
    \item \textbf{One-step-ahead forecasting is enough.} Motion prediction, specifically for MOT, requires only the immediate predicted position in the current timestamp to be matched with detections. At each time step, the predicted position will be assessed and refined based on the detection information from the external detector. The matched detections will be stored and used as the trajectory of the given object rather than the predicted bounding boxes from the motion predictor. Therefore, the need for generative ability or to predict multiple steps simultaneously is limited. Moreover, the complex and computationally expensive "feature" of the generative or autoregressive model is unnecessary and overkill.
    
    \item \textbf{Preventing Autoregressive Drift.} The potential cumulative error occurring during the autoregressive process can lead to poor performance over time, which is hard to overcome, especially in noisy situations using detections from an external detection model like the tracking-by-detection paradigm. 
\end{itemize}

Our proposed model leverages the TCN in a direct manner: The TCN captures temporal dependencies and models non-linear motion dynamics, transforming raw temporal inputs into rich motion features. These features are then passed to the Regression Head, which predicts the motion $\mathbf{M}_{m+1}$ in a single step. This design balances simplicity and computational efficiency while ensuring superior performance in motion prediction tasks for MOT.

\subsubsection{Modified Temporal Convolutional Network}

Our TCN is a series of $K$ identical blocks, where the output of each block is the input to the next block. Each block takes an initial prediction of the prior block and refines it. The only exception is the first block, which has no prior block; thus, its input is the overall input, namely $\mathbf{C}_{m}$. 

Each block is a stack of $L$ layers, $E^{(l)} \in \mathbb{R}^{F_l \times m}$, where $F_l$ is defined as the number of the convolutional kernels in the $l$ layer and $m$ is the input context length. In each layer, $l$, the input is passed through a series of operations, including a dilated causal convolution operation, an activation function, and a dropout. The dilation rate increases for consecutive layers within a block, such that the layer $l$-th will have the dilation rate $d_l = 2^{l-1}$. Formally, with reference to Equation \ref{eq:dilated-conv}, the intermediate hidden feature at time $t$ of layer $l$-th in block $k$-th, denoted as $h_{t}^{k, l}$ is calculated as:
\begin{align}
        h_{t}^{k, l} & = \displaystyle \phi_l \left[ (h_{t}^{k, l-1} \ast_{d_l} W^l)(t) \right]  \\
                     & = \displaystyle \phi_l \left[ \sum_{a = s + d_{l}t}{h_{t-a}^{k, l-1}(a) \times W^{k,l}(t-a)} \right]
\end{align}

where $\ast_{d_l}$ is the dilated convolution operation, $W^{k,l}$ is the parameterized convolution kernel at layer $l$, $\phi(\cdot)$ is the activation function. We adopt the gated activation unit mechanism \cite{van2016pixelcnn}, as follows:
\begin{equation}
    \phi_L(h) = \tanh(W_f^{k, l} \ast_{d_l} h) \odot \sigma (W_g^{k, l} \ast_{d_l} h) ,
\end{equation}

where $\odot$ is an element-wise multiplication operation, $\tanh(\cdot)$ is the tanh activation function, $\sigma(\cdot)$ is the sigmoid function, $W_f^{k, l}$ is the "filter" convolution kernel in layer $l$ of block $k$, and $W_g^{k, l}$ is the "gate" convolution kernel in layer $l$ block $k$. Let $\hat{z}^{k, l}$ and $z^{k, l}$ be the skipped connection from layer $l$ of block $k$ and the result after layer $l$ of block $k$, respectively. 
\begin{align}
    \label{eq:output-of_block}
    \hat{z}^{k, l} &= \left(\phi_L \left(h^{k, l-1}\right) \ast W^{k, l}_{\text{skip}} \right) \\
    z^{k, l} &= \frac{1}{\sqrt{2}} \left[ \phi_L \left(h^{k, l-1}\right) \ast W^{k, l}_{\text{out}} + z^{k, l-1} \right]
\end{align}

where $\frac{1}{\sqrt{2}}$ is the normalization term, $W^{k, l}_{\text{skip}}$ and $W^{k, l}_{\text{out}}$ are two $1 \times 1$ parameterized convolution kernel of block $k$, and $z^{k-1}$ is the output from previous block. We define $z^{0} = \mathbf{C}_{m}$ so that Equation \ref{eq:output-of_block} will be valid for all $1 \le k \le K$.

Although TCN can work with any arbitrary context length, we restrain it to using the max length of $T = 16$. This aligns well with the receptive field of our architecture, which has a kernel size of $F=2$ and a number of layers $L=4$. This decision is based on empirical experiments and is discussed in detail in Section \ref{sec:ablation-study}. Note that our model retains the well-utilized ability to work with trajectories of any length.

\subsection{Regression Head}

After data is passed through the TCN, we now have 2 sources of information: the parameterized skipped connections $\{ \hat{z}^{1}, \dots, \hat{z}^{L} \}$ and the output from the final block $z^K$. We aggregate all skipped connections to create a hierarchical representation $\hat{z}$ as follows:
\begin{equation}
    \hat{z}_i = \sum_j^L{\hat{z}^{j}_{i}} \quad \forall i \in \{1, \dots, L\}
\end{equation}

Which results in a feature of the same size as $z^K$. To utilize $\hat{z}$ and $z^K$ without explicitly indicating their significance, we introduce a new learned parameter $\alpha \in \mathbb{R}$ that can be trained alongside our model. The combined feature $\textbf{z} \in \mathbb{R}^{F_L \times m}$ is calculated as:
\begin{equation}
    \label{eq:alpha}
    \textbf{z} = \alpha z^K + (1 - \alpha) \hat{z}
\end{equation}

We find that the $\alpha$ varies across different scenarios. Specifically, in the trained model reported in this paper, $\alpha = 0.557$ and $\alpha = 0.482$ on DanceTrack and SportsMOT, respectively. This difference in the $\alpha$ value implies that the contribution of $z^K$ and $\hat{z}$ differ depending on the scenario, and the $\alpha$ parameter can dynamically adapt to these variations.

Global average pooling (GAP) is employed to aggregate features across the temporal dimension of $\textbf{z}$, resulting in a compact representation for each channel. Specifically, with $F_L = 4$, the result of GAP of $\textbf{z}$ will be the predicted motion $\widehat{\mathbf{M}}_{m+1}$, such that:
\begin{equation}
    \widehat{\mathbf{M}}^{\{f\}}_{m+1} = \frac{1}{m} \sum_{t=1}^{m} \mathbf{z}_{f, t} \quad \forall f \in \{1, \dots, F_L\}
\end{equation}

\subsection{Loss function}

We use the $L_2$ (Mean Squared Error) as our Loss function. Given the ground truth $\mathbf{M}_{m+1}$ and the predicted motion $\widehat{\mathbf{M}}_{m+1}$, the MSE loss is calculated as follows:

\begin{equation}
    L = L_2(\mathbf{M}_{m+1}, \widehat{\mathbf{M}}_{m+1})  = \sum{(\mathbf{M}_{m+1} - \widehat{\mathbf{M}}_{m+1})^2}
\end{equation}

In our experiments, models trained with $L_2$ loss have a more stable training process and perform better overall compared to models trained with smooth $L_1$.

\subsection{Data Augmentation}
\label{sec:data-augmentation}

We introduced a new data augmentation process for training the TCMP to utilize the capability of handling contexts of arbitrary length and noisy inputs.
 
\textbf{Add noises}. We strategically add noise into training $\mathbf{C}_{m}$ by first sampling a noise vector $\mathbf{N} \in \mathbb{R}^{B \times m \times 4}$ with $B$ is the batch size, $m$ is the max context length and $4$ is for $(x, y, w, h)$, respectively. The noise vector $\mathbf{N}$ is sampled from a normal distribution $\mathbf{N}_i \sim \mathcal{N}_4 (\mu, \sigma)$ where $\mu \in \mathbb{R}^{4}$ and $\sigma \in \mathbb{R}^{4 \times 4}$. Specifically, we use $\mu = 0$ and $\sigma = 0.001$ in our training process, as this is the empirical basic for setting noise value. After being sampled, $\mathbf{N}$ will be add to $\mathbf{C}_{m}$ as follows:
\begin{equation}
    \mathbf{C}_{t}^{(b, i)} = \mathbf{C}_{t}^{(b, i)} + \mathbf{N}^{b, t, i} \quad \forall b \in \{1, \dots, B\}, \forall i \in \{1, \dots, 4\}, \forall t \in \{1, \dots, m\}
\end{equation}  

Where $\mathbf{C}_{t}^{(b, i)}$ is the value $i$-th of time $t$ in batch $b$. This operation makes the motion information (last $4$ value in $\mathbf{C}_i$) misaligned; we recalculate it using Equation \ref{eq:e_m_ot} and Equation \ref{eq:c_t}. This strategy simulates the noisy and unreliable detections from the external detectors, making the trained model robust and less sensitive to noises.

\textbf{Random Context Lengths}. To train the model with varying context lengths, we randomly truncate the contexts in batches after they are created with a fixed context length of $m$. We sample a random number $m'$ from a uniform distribution, $m' \sim \mathcal{U}_{[4, m]}$, and only use $m'$ latest contexts instead of $m$. Thus, the new context is now become:
\begin{equation}
    \mathbf{C}' = \{\mathbf{C}_t\} \quad \forall t \in \{ m - m', \dots, m \}
\end{equation}

With this setup, contexts will have a minimum length of $4$ and a maximum length of $m$.

\section{Experiments}

\subsection{Datasets}

We performed experiments on two well-known complex motion datasets, namely DanceTrack\cite{sun2022dancetrack} and SportsMOT\cite{cui2023sportsmot}. Both datasets were recently proposed for multi-object tracking, specifically on human tracking, with complex motion patterns and uniform appearances. The objects in both datasets are frequently occluded and crossovers, raising a significant challenge to current methods that assume linear motion. The DanceTrack dataset emphasizes uniform appearance and diverse body gestures, while the main challenge of SportsMOT arises from high-speed camera movement and occlusion.

\subsection{Evaluation Metrics}

We evaluate our method on Higher Order Metric (HOTA, AssA, DetA) \cite{luitenHOTA2021}, IDF1 \cite{ristaniIDF12016}, and MOTA from CLEAR metrics \cite{bernardinCLEAR2008}. HOTA is a well-adopted metric that explicitly balances the effects of performing accurate detection and association. IDF1 and AssA are used for association performance evaluation. DetA and MOTA primarily evaluate detection performance. Since the ultimate goal of a motion predictor in MOT is to improve the association step, the improvements in HOTA, IDF1, and AssA are more significant. 

\subsection{Experimental Procedure}

\begin{figure*}[h]
    \centering
    \includegraphics[width=1\linewidth]{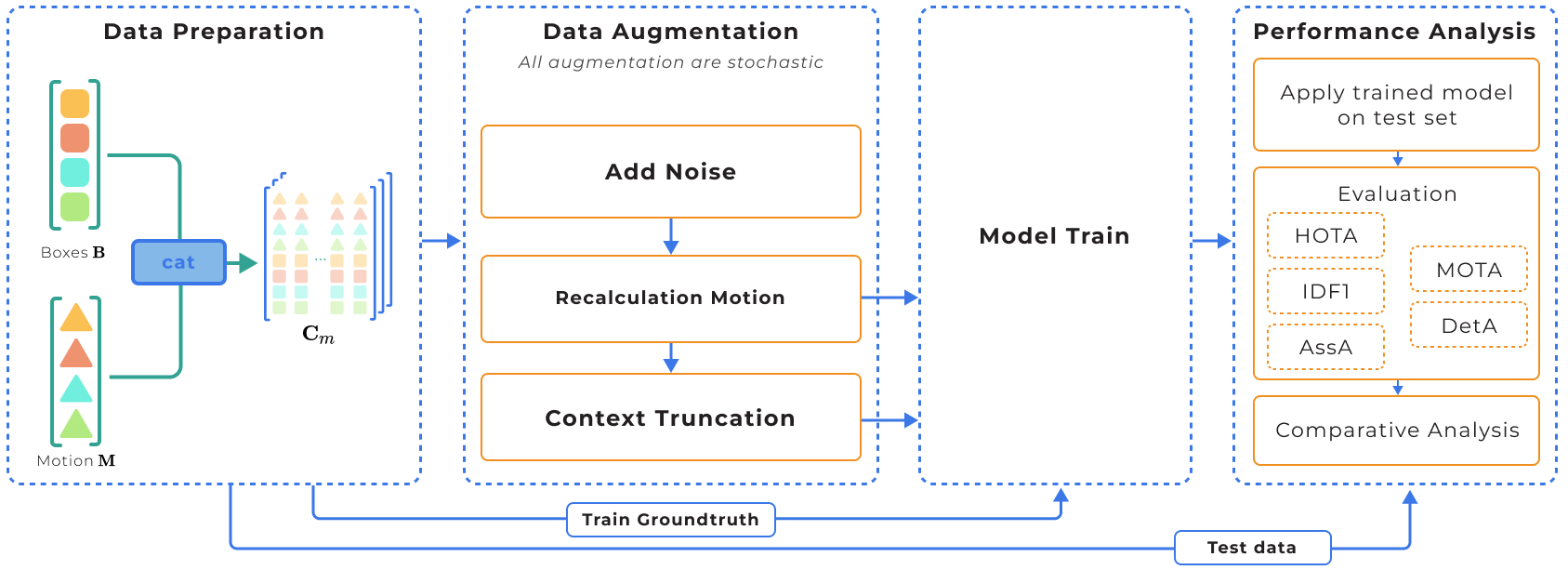}
    \caption{The detail of our experimental procedure}
    \label{fig:experimental-procedure}
\end{figure*}

Our experimental procedure, illustrated in Figure \ref{fig:experimental-procedure}, follows a comprehensive sequence of data preparation, data augmentation, model training, and performance analysis. 

\begin{itemize}
    \item \textbf{Data Preparation}. The initial stage of the experimental procedure involves preparing the data for model training. This process follows our previously defined problem in Section \ref{sec:problem-statement}:  bounding boxes $\mathbf{B}_T$ are first used to calculate motion information $\mathbf{M}_T$, then $\mathbf{B}_T$ and $\mathbf{M}_T$ are concatenated (column-wise) to form $\mathbf{C}_T$. This calculation is done for every boxes in every batch. The output of this stage is a structured context information $\mathbf{C}_m$ ready for augmentation.

    \item \textbf{Data Augmentation}. The prepared data is then subjected to a series of stochastic data augmentation techniques.  These augmentations are crucial for improving the model's robustness and generalization ability.
The detail information of our data augmentation process is discussed in Section \ref{sec:data-augmentation}

    \item \textbf{Model Train}. The augmented dataset and the train groundtruth (the real object's motion) are used as input for the model training phase. Following this iterative process, the trained model is utilized to conduct a performance analysis.

    \item \textbf{Performance Analysis}. The final stage involves a thorough performance analysis of the trained model. The model's motion prediction ability is evaluated on test sets on standard multi-object tracking metrics, including HOTA, MOTA, IDF1, AssA, and DetA. These metrics quantify different aspects of tracking performance, including detection accuracy, identity preservation, and association accuracy.  Finally, a comparative analysis compares the model's performance against other state-of-the-art methods on the same benchmark datasets. This provides a comprehensive assessment of the model's effectiveness.
\end{itemize}

\subsection{Implementation Details}

We primarily focus on developing a motion model for tracking objects; thus, for a fair comparison, we use the publicly available YOLOX detector weights provided by ByteTrack \cite{zhangByteTrack2022}, DanceTrack \cite{sunDanceTrack2022} and SportsMOT \cite{cuiSportsMOT2023} to detect objects in the MOT, DanceTrack and SportsMOT datasets, respectively.

The detailed design of our proposed model is shown in Table \ref{tab:model_detail_architecture}. Our baseline uses a stack of $K=2$; each block contains $L=4$ layer of dilated convolution layers, where the dilation factor is doubled at each layer; specifically, the dilation factors are $[1, 2, 4]$. All layers use $64$ $2 \times 2$ filter. We utilize layer normalization \cite{ba2016layer} and weight normalization \cite{salimans2016weight} techniques to stabilize training. In all experiments, we use a batch size of 1024, Adam optimization technique \cite{kingma2017adammethodstochasticoptimization} with $\beta_1 = 0.9$, $\beta_2 = 0.999$, and learning rate of $2.5e^{-4}$ with no weight decay.

 For the association stage, we adopt the BYTE association method, which performs separate matching for high-scoring and low-scoring boxes. The high threshold $\tau_{\text{high}} = 0.6$ and the low threshold $\tau_{\text{low}} = 0.4$. We only use appearance similarity in the first association. The detailed algorithm and further discussion about the association stage are in Appendix \ref{sec:pseudo-code}.

\begin{table}[t!]
\centering
\caption{Detail layers design of our TCMP proposed architecture. $m$ is the context $\mathbf{C}_{m}$'s length, $K$ is the number of block, $L$ is the number of layer}
\label{tab:model_detail_architecture}
    \begin{tabular}{llcc}
    \toprule
    \multicolumn{1}{c}{\textbf{Layer}}               & \multicolumn{1}{c}{\textbf{Type}}         & \textbf{Input Shape} & \textbf{Output Shape} \\ \midrule
    \multicolumn{4}{c}{\textbf{Input}} \\ \midrule

    Input Conv                   & Conv1d           & $8 \times m$         & $64 \times m$         \\ \midrule
    Layer Norm                   & LayerNorm              & $64 \times m$        & $64 \times m$         \\ \midrule
    ReLU                         & ReLU                   & $64 \times m$        & $64 \times m$         \\ \midrule
    Dropout                      & Dropout ($p=0.2$)        & $64 \times m$        & $64 \times m$         \\ \midrule
    \multicolumn{4}{c}{
        \begin{tabular}{@{}c@{}}
            \textbf{Layer} \\  
            ($K \times L = 8$)
        \end{tabular}
    }
    \\ \midrule
    \multirow{8}{*}{Layer $i$}   & Stacked Dilated Conv1d & $64 \times m$       & $64 \times m$        \\ \cmidrule(l){2-4} 
                                 & Residual Connection    & $64 \times m$       & $64 \times m$         \\ \cmidrule(l){2-4} 
                                 & Skip Connection        & $64 \times m$       & $64 \times m$         \\ \cmidrule(l){2-4} 
                                 & "Filter" Conv          & $64 \times m$       & $64 \times m$         \\ \cmidrule(l){2-4} 
                                 & "Gate" Conv            & $64 \times m$       & $64 \times m$         \\ \cmidrule(l){2-4} 
                                 & LayerNorm              & $64 \times m$       & $64 \times m$         \\ \cmidrule(l){2-4} 
                                 & ReLU                   & $64 \times m$       & $64 \times m$         \\ \cmidrule(l){2-4} 
                                 & Dropout ($p=0.2$)      & $64 \times m$       & $64 \times m$         \\ \midrule
    \multicolumn{4}{c}{\textbf{Regression Head}} \\ \midrule
    ReLU                         & ReLU                   & $64 \times m$       & $64 \times m$         \\ \midrule
    Output Conv 1                & Conv1d                 & $64 \times m$       & $64 \times m$         \\ \midrule
    ReLU                         & ReLU                   & $64 \times m$       & $64 \times m$         \\ \midrule
    Output Conv 2                & Conv1d                 & $64 \times m$       & $4  \times m$         \\ \midrule
    GAP                          & Global Average Pooling & $4  \times m$       & $4  \times 1$         \\ \bottomrule
    \end{tabular}
\end{table}

\subsection{Benchmark Evaluation}
\label{sec:benchmark-evaluation}

\begin{table}[t!]
    \centering
    \caption{Comparison of the state-of-the-art trackers on DanceTrack test set. All trackers are trained without extra training data. \colorbox{sky}{Trackers in the blue block} share the same detections. $\uparrow$ means the higher the better. \textbf{Bold} numbers indicate the best result.}
    \label{tab:exp-results-dancetrack}
        \begin{tabular}{l|rrrrr}
            \toprule
            \textbf{Tracker} & \textbf{HOTA} $\uparrow$ & \textbf{IDF1} $\uparrow$ & \textbf{AssA} $\uparrow$ & \textbf{MOTA} $\uparrow$ & \textbf{DetA} $\uparrow$ \\
            \midrule
            FairMOT \cite{zhangFairMOT2021} & 39.7 & 40.8 & 23.8 & 82.2 & 66.7 \\
            CenterTrack \cite{zhouCenterTrack2020} & 41.8 & 35.7 & 22.6 & 86.8 & 78.1 \\
            TraDes \cite{wuTraDes2021} & 43.3 & 41.2 & 25.4 & 86.2 & 74.5 \\
            TransTrack \cite{sunTransTrack2020} & 45.5 & 45.2 & 27.5 & 88.4 & 75.9 \\
            QDTrack \cite{fischer2023qdtrack} & 45.7 & 44.8 & 29.2 & 83.0 & 72.1 \\
            DiffusionTrack \cite{luo2024diffusiontrack} & 52.4 & 47.5 & 33.5 & 89.5 & 82.2 \\
            MOTR \cite{zengMOTR2022} & 54.2 & 51.5 & 40.2 & 79.7 & 73.5 \\
            \rowcolor{sky} DeepSORT \cite{wojkeDeepSORT2017} & 45.6 & 47.9 & 29.7 & 87.8 & 71.0 \\
            \rowcolor{sky} ByteTrack \cite{zhangByteTrack2022} & 47.3 & 52.5 & 31.4 & 89.5 & 71.6 \\
            \rowcolor{sky} SORT \cite{bewleySORT2016} & 47.9 & 50.8 & 31.2 & 91.8 & 72.0 \\
            \rowcolor{sky} OC-SORT \cite{caoOCSort2023} & 55.1 & 54.2 & 38.0 & 89.4 & 80.3 \\
            \rowcolor{sky} StrongSORT \cite{du2023strongsort} & 55.6 & 55.2 & 38.6 & 91.1 & 80.7 \\
            \rowcolor{sky} SparseTrack \cite{liu2025sparsetrack} & 55.5 & 58.3 & 39.1 & 91.3 & 78.9 \\
            \rowcolor{sky} MotionTrack \cite{xiao2024motiontrack} & 58.2 & 58.6 & 41.7 & 91.3 & 81.4 \\
            \rowcolor{sky} C-BIoU \cite{yang2023hard} & 60.6 & 61.6 & 45.4 & 91.6 & 81.3 \\
            \rowcolor{sky} Deep OC-SORT \cite{caoOCSort2023} & 61.3 & 61.5 & 45.8 & 92.3 & 82.2 \\
            \rowcolor{sky} DiffMOT \cite{lv2024diffmot} & 62.3 & 63.0 & 47.2 & \textbf{92.8} & \textbf{82.5} \\
            \rowcolor{sky} Ours & \textbf{63.4} & \textbf{65.0} & \textbf{49.1} & 92.0 & 82.0 \\
            
            \bottomrule
        \end{tabular}
\end{table}

\subsubsection{DanceTrack}

To evaluate TCMP under challenging non-linear object motion, we report results on the DanceTrack in Table \ref{tab:exp-results-dancetrack}. Our proposed method significantly improves upon the previous best method, DiffMOT, in several key metrics: HOTA increases from 62.3\% to 63.4\%, IDF1 rises from 63.0\% to 65.0\%, and AssA improves from 47.2\% to 49.1\%.
As shown in Table \ref{tab:exp-results-dancetrack}, TCMP not only surpasses the recent DiffMOT but also achieves substantial gains over widely-used trackers such as ByteTrack (+16.1\% HOTA) and SORT (+15.5\% HOTA).
Significantly, TCMP achieves these gains in accuracy while maintaining exceptional efficiency, as it is only 0.014 times the size (in terms of parameters) and requires only 0.05 times the computational cost (in terms of FLOPs) compared to DiffMOT. This combination of high accuracy and low computational overhead demonstrates TCMP's effectiveness and practicality for real-world applications involving complex object motion. See Figure \ref{fig:qualitative-dancetrack} for qualitative results. 

Furthermore, as detailed in our ablation study (Section \ref{sec:ablation-study-motion-models}), our approach significantly outperforms classic baselines, such as the Kalman Filter and simple IoU matching, demonstrating the value of our learned non-linear motion model.

\subsubsection{SportsMOT}

Results on SportsMOT are shown in Table \ref{tab:exp-results-sportsmot}. SportsMOT is another challenging dataset specifically designed to evaluate tracking performance under conditions of frequent and complex non-linear motion in sports. In SportsMOT, there are two settings for testing performance, each utilizing a different detector. 

In the first setting (using a detector trained only on the SportsMOT train set), TCMP achieves a HOTA score of 73.3\%, an IDF1 score of 74.2\%, and an AssA score of 62.6\%. Compared to previous SOTA, TCMP improves HOTA by 1.2 percentage points (from 72.1\% to 73.3\%), IDF1 by 1.4 percentage points (from 72.8\% to 74.2\%), and AssA by 2.1 percentage points (from 60.5\% to 62.6\%). 

Compared more broadly against other methods evaluated under this setting (Table \ref{tab:exp-results-sportsmot}), TCMP not only surpasses the prior state-of-the-art DiffMOT but also demonstrates notable advantages over other competitive trackers like OC-SORT, BOT-SORT, and ByteTrack, particularly showcasing superior performance in the crucial association metrics HOTA, IDF1 and AssA. This highlights TCMP's effectiveness in handling the complex sports scenarios even with this detector setup. Since the goal of a motion predictor in MOT is to improve the association step, the improvements in HOTA, IDF1, and AssA are more significant. 

TCMP continues to demonstrate strong performance under the second setting (using a detector trained on the SportsMOT train and val sets). TCMP firmly outperformed all other motion-based trackers. It achieves state-of-the-art results with 76.3\% HOTA, 76.5\% IDF1, and 65.3\% AssA. This performance not only surpasses the previous SOTA method, DiffMOT, across all key metrics (HOTA +0.1, IDF1 +0.4, AssA +0.2), but also significantly outperforms other widely-recognized methods like ByteTrack, BoTSORT, and OC-SORT by a substantial margin. The consistent improvements show TCMP's superior ability to predict objects' motion with complex, non-linear scenarios, even under the variations present between the two SportsMOT settings.

\begin{table}[t!]
    \centering
    \caption{Comparison of the state-of-the-art trackers on SportsMOT test set. \colorbox{sky}{Trackers in the blue block} share the same detections. Trackers with * indicate that their detectors are trained on SportsMOT train and val sets. $\uparrow$ means the higher the better. \textbf{Bold} numbers indicate the best result.}
    \label{tab:exp-results-sportsmot}
    \begin{tabular}{l|rrrrr}
        \toprule
        \textbf{Tracker} & \textbf{HOTA} $\uparrow$ & \textbf{IDF1} $\uparrow$ & \textbf{AssA} $\uparrow$ & \textbf{MOTA} $\uparrow$ & \textbf{DetA} $\uparrow$ \\
        \midrule
        FairMOT \cite{zhangFairMOT2021} & 49.3 & 53.5 & 34.7 & 86.4 & 70.2 \\
        GTR \cite{zhou2022global} & 54.5 & 55.8 & 45.9 & 67.9 & 64.8 \\
        QDTrack \cite{fischer2023qdtrack} & 60.4 & 62.3 & 47.2 & 90.1 & 77.5 \\
        CenterTrack \cite{zhouCenterTrack2020} & 62.7 & 60.0 & 48.0 & 90.8 & 82.1 \\
        TransTrack \cite{sunTransTrack2020} & 68.9 & 71.5 & 57.5 & 92.6 & 82.7 \\
        \rowcolor{sky} ByteTrack \cite{zhangByteTrack2022} & 62.8 & 69.8 & 51.2 & 94.1 & 77.1 \\
        \rowcolor{sky} BoT-SORT \cite{aharonBoTSORT2022} & 68.7 & 70.0 & 55.9 & 94.5 & 84.4 \\
        \rowcolor{sky} OC-SORT \cite{caoOCSort2023} & 71.9 & 72.2 & 59.8 & 94.5 & \textbf{86.4} \\
        \rowcolor{sky} DiffMOT \cite{lv2024diffmot} & 72.1 & 72.8 & 60.5 & 94.5 & 86.0 \\
        \rowcolor{sky} Ours & \textbf{73.3} & \textbf{74.2} & \textbf{62.6} & 94.1 & 86.0 \\
        \midrule
        \rowcolor{sky} ByteTrack* \cite{zhangByteTrack2022} & 64.1 & 71.4 & 52.3 & 95.9 & 78.5 \\
        \rowcolor{sky} MixSort-Byte* \cite{cuiSportsMOT2023} & 65.7 & 74.1 & 54.8 & 96.2 & 78.8 \\
        \rowcolor{sky} OC-SORT* \cite{caoOCSort2023} & 73.7 & 74.0 & 61.5 & 96.5 & 88.5 \\
        \rowcolor{sky} MotionTrack* \cite{xiao2024motiontrack} & 74.0 & 74.0 & 61.7 & 96.6 & 88.8 \\
        \rowcolor{sky} MixSort-OC* \cite{cuiSportsMOT2023} & 74.1 & 74.4 & 62.0 & 96.5 & 88.5 \\
        \rowcolor{sky} DiffMOT* \cite{lv2024diffmot} & 76.2 & 76.1 & 65.1 & \textbf{97.1} & \textbf{89.3}\\
        \rowcolor{sky} Ours* & \textbf{76.3} & \textbf{76.5} & \textbf{65.3} & 96.8 & 89.2 \\
        \bottomrule
    \end{tabular}
\end{table}

\subsection{Ablation Studies}
\label{sec:ablation-study}

We conduct a rigorous series of experiments (ablation studies) on the validation set of DanceTrack with the same YOLOX-X detector throughout all experiments to ensure the consistency. In general, the ablation studies highlight the importance of both the skipped connections and the final block output in the TCMP architecture, with the parameterized $\alpha$ effectively balancing their contributions (Section \ref{sec:ablation-study-alpha}). The nature of the temporal convolutional network makes it work with arbitrary length context length, and from detailed experiments, we discover that TCMP works best on context length up to 16; longer context only adds outdated information and does not improve the performance (Section \ref{sec:ablation-study-context-length}). On comparing with other Motion models, our proposed TCMP outperforms other methods by a significant margin, illustrating the robustness of the proposed methods (Section \ref{sec:ablation-study-motion-models}). While achieving state-of-the-art performance, our proposed TCMP, compared to the prior SOTA method, is only $0.014$ times the parameter size and $0.05$ times the computational cost (Section \ref{sec:ablation-study-flops}). Finally, TCMP maintains high performance when trained and tested on different datasets, indicating its ability to generalize well on different scenarios (Section \ref{sec:ablation-study-generalization}). 

\subsubsection{Impact of the parameterized $\alpha$}
\label{sec:ablation-study-alpha}

To further understand and prove the influence of the parameterized $\alpha$, we assess the regression performance under a detailed ablation study with three distinct configurations of our model's architecture:

\begin{enumerate}
    \item \textbf{Using only the parameterized skipped connections ($\hat{z}$ only)}. In this setup, the regression head utilizes information solely from the parameterized skipped connections, bypassing the contribution of the TCN's final output. This configuration isolates the impact of features directly passed through the skipped connections, emphasizing the raw and early-stage temporal information.

    \item \textbf{Using only the output from the final block ($z^K$ only)}. We examine the performance when the regression head relies entirely on the processed features from the TCN's final block, $z^K$. This setup allows us to analyze the significance of the deeper, hierarchically structured features extracted by the TCN, which capture the refined temporal patterns over the sequence.

    \item \textbf{Using a weighted combination of both ($\alpha \hat{z} + (1 - \alpha) z^K$)}. In the most comprehensive configuration, both the skipped connections and the final block output are combined using a weighted parameter $\alpha$ to control their relative contributions. This approach leverages both early-stage and refined features, creating a synergy between them.
\end{enumerate}

The results of these experiments are presented in Table \ref{tab:ablation-alpha}. The outcomes reveal that both $z^K$ and $\hat{z}$ contribute valuable information for motion prediction. Precisely, $z^K$ captures global temporal dependencies and refined patterns that are critical for accurate motion modeling, while $\hat{z}$ retains localized and raw temporal information that can complement the final output. The weighted combination controlled by $\alpha$ demonstrates the best performance, proving that the contribution of $\alpha$ in our design is significant and necessary.

\begin{table}[t!]
    \centering
    \caption{Ablation study results for the parameterized $\alpha$. The table compares regression performance when using only skipped connections, only the final block output, and their weighted combination, demonstrating the impact of $\alpha$ on the model's effectiveness.}
    \label{tab:ablation-alpha}
    \begin{tabular}{l|rrrrr}
        \toprule
        \textbf{Configuration} & \textbf{HOTA} $\uparrow$ & \textbf{IDF1} $\uparrow$ & \textbf{AssA} $\uparrow$ & \textbf{MOTA} $\uparrow$ & \textbf{DetA} $\uparrow$ \\
        \midrule
        $z^K$ only                         & 58.3          & 58.7          & 43.2          & 89.9 & 79.1 \\
        $\hat{z}$ only                     & 57.7          & 58.0          & 42.2          & 89.7 & \textbf{79.3} \\
        $\alpha z^K + (1 - \alpha)\hat{z}$ & \textbf{58.8} & \textbf{60.0} & \textbf{43.9} & \textbf{90.0} & 79.2 \\
        \bottomrule
    \end{tabular}
\end{table}

\subsubsection{Impact of different Context lengths}
\label{sec:ablation-study-context-length}

Although our model can work with arbitrary length of input context, we are curious how our model performs on different context lengths. We perform the experiments by limiting our model to work with context lengths up to a certain number. The results are shown in Table \ref{tab:ablation-different-context-length}. The results from this ablation study explain why our model is constrained on a fixed max context length of $m = 16$. 

\begin{table}[t!]
    \centering
    \caption{Ablation study results for different context length.}
    \label{tab:ablation-different-context-length}
    \begin{tabular}{l|rrrrr}
        \toprule
        \begin{tabular}{@{}c@{}}\textbf{Context length} \\ $m$\end{tabular} & \textbf{HOTA} $\uparrow$ & \textbf{IDF1} $\uparrow$ & \textbf{AssA} $\uparrow$ & \textbf{MOTA} $\uparrow$ & \textbf{DetA} $\uparrow$ \\
        \midrule
        $m \le 4$   & 57.0          & 57.5          & 41.3          & 89.8          & 79.2          \\
        $m \le 8$   & 57.3          & 58.0          & 41.9          & 89.8          & 78.8          \\
        $m \le 16$  & \textbf{58.8} & \textbf{60.0} & \textbf{43.9} & \textbf{90.0} & 79.2          \\
        $m \le 24$  & 58.6          & 58.0          & 43.6          & 89.9          & 79.5          \\
        $m \le 32$  & 57.8          & 57.5          & 42.2          & 89.8          & \textbf{79.4} \\
        \bottomrule
    \end{tabular}
\end{table}

As shown in Table \ref{tab:ablation-different-context-length}, as the context length increases beyond a certain point, the performance of the model begins to plateau. The minor differences in the results suggest that TCMP can effectively capture relevant information on a wide range of context lengths, and there is an optimal max context length that motion predictors work best on, empirically, $m = 16$. 

 While longer contexts might contain richer information, they also introduce challenges such as increased computational requirements and potential information overload, which can lead to diminished performance. In addition, the motion in DanceTrack and SportsMOT are highly complex, and the changes in motion in each frame are subtle and fast; longer contexts, as a result, may not contain relevant information.

 Our findings indicate that while the capability to work with arbitrary context length is advantageous, a fixed maximum context length can yield consistent and reliable performance.

\subsubsection{Different motion models}
\label{sec:ablation-study-motion-models}

To evaluate the effectiveness of our proposed motion model in non-linear motion scenarios, we conducted experiments comparing it with various existing motion models; the results are shown in Table \ref{tab:ablation-different-motion-model}. We compare various baseline models, including no motion model (IoU only), the linear Kalman Filter (KF), and non-linear motion models, namely LSTM, Transformer-based and diffusion-based approaches. Our method consistently outperforms others, achieving the highest scores: 58.8\% in HOTA, 60.0\% in IDF1, 43.9\% in AssA, 90.0\% in MOTA, and 79.2\% in DetA. This result highlights the robustness of our designed TCN in non-linear motion predictions, as it directly learns the distribution of all objects' motions across the dataset rather than focusing on individual trajectories. Compared to the Diffusion-based method, our model delivers up to 3.1\% and 4.8\% improvements in HOTA and IDF1, respectively. These results, along with results in Table \ref{tab:exp-results-dancetrack}, underscore the capability of our model to deal with complex motion scenarios.

 \begin{table}[t!]
    \centering
    \caption{Comparison of different motion models on the DanceTrack validation sets. The best results are shown in \textbf{bold}.}
    \label{tab:ablation-different-motion-model}
    \begin{tabular}{l|rrrrr}
        \toprule
        \textbf{Method} & \textbf{HOTA} $\uparrow$ & \textbf{IDF1} $\uparrow$ & \textbf{AssA} $\uparrow$ & \textbf{MOTA} $\uparrow$ & \textbf{DetA} $\uparrow$ \\
        \midrule
        IoU only & 44.7 & 36.8 & 25.3 & 87.3 & \textbf{79.6} \\
        Kalman Filter & 46.8 & 52.1 & 31.3 & 87.5 & 70.2 \\
        LSTM & 51.2 & 51.6 & 34.3 & 87.1 & 76.7 \\ 
        Transformer & 54.6 & 54.6 & 38.1 & 89.2 & 78.6 \\ 
        Diffision & 55.7 & 55.2 & 39.5 & 89.3 & 78.9 \\
        TCN (Ours) & \textbf{58.8} & \textbf{60.0} & \textbf{43.9} & \textbf{90.0} & 79.2 \\
        \bottomrule
    \end{tabular}
\end{table}

\subsubsection{The number of Parameters and FLOPs}
\label{sec:ablation-study-flops}

Table \ref{tab:exp-params-and-flops} shows the comparison of our proposed motion model with prior state-of-the-art motion models, D$^2$MP, in terms of parameters, FLOPs, and HOTA scores on the DanceTrack and SportsMOT datasets. FLOPs is calculated with $X \in \mathbb{R}^{1 \times 5 \times 8}$, representing batch size, context length $M$ and context size (Equation \ref{eq:c_t}), respectively. 

TCMP achieves a significant reduction in both parameters (0.207M) and FLOPs (1.526M), which is $0.014$ times the size and $0.05$ times the computational cost of D$^2$MP. Despite its simplicity, our method achieves the highest HOTA scores on both datasets, surpassing D$^2$MP by $+1.1\%$ on DanceTrack and a slight $+0.1\%$ on SportsMOT. These results highlight the model's ability to maintain state-of-the-art performance while significantly reducing computational complexity, setting a new standard for efficient and accurate multi-object tracking.  

\begin{table}[t!]
    \centering
    \caption{Comparison of the number of model's parameters (\# Params) and Floating Point Operations (FLOPs), with the corresponding HOTA on DanceTrack (details in Table \ref{tab:exp-results-dancetrack}) and SportsMOT (details in Table \ref{tab:exp-results-sportsmot}).}
    \label{tab:exp-params-and-flops}
    \begin{tabular}{l|rr|rr}
        \toprule
        \textbf{Tracker} & \textbf{\# Params} & \textbf{FLOPs} $\downarrow$ & \begin{tabular}{@{}c@{}}\textbf{HOTA} $\uparrow$ \\ \textit{DanceTrack}\end{tabular} & \begin{tabular}{@{}c@{}}\textbf{HOTA} $\uparrow$ \\ \textit{SportsMOT}\end{tabular} \\ \midrule
        % MotionTrack \cite{xiao2024motiontrack} & - & - & 58.2 & 74.0\\
        D$^2$MP \cite{lv2024diffmot} & 14.7M & 28.0M & 62.3 & 76.2\\
        TCMP (Our) & 0.2M & \textbf{1.5M} & \textbf{63.4} & \textbf{76.3} \\
        \bottomrule
    \end{tabular}
\end{table}

\subsubsection{Generalization of TCMP}
\label{sec:ablation-study-generalization}

 \begin{table}
    \centering
    \caption{Generalization experiment of TCMP.}
    \label{tab:ablation-generalization}
    \begin{tabular}{ll|rrrrr}
    \toprule
    \textbf{Test} & \textbf{Train} & \textbf{HOTA} $\uparrow$ & \textbf{IDF1} $\uparrow$ & \textbf{AssA} $\uparrow$ & \textbf{MOTA} $\uparrow$ & \textbf{DetA} $\uparrow$ \\
    \midrule
    \multirow{2}{*}{DanceTrack} & DanceTrack & 58.8 & 60.0 & 43.9 & 90.0 & 79.2 \\
                                & SportsMOT  & 58.3 & 58.7 & 43.2 & 89.9 & 79.1 \\
                                \midrule
    \multirow{2}{*}{SportsMOT}  & SportsMOT  & 83.4 & 83.1 & 74.2 & 98.5 & 93.8 \\
                                & DanceTrack & 82.0 & 81.3 & 72.1 & 97.9 & 93.4 \\
    \bottomrule
    \end{tabular}
\end{table}

We systematically train and test TCMP on 2 datasets: DanceTrack, and SportsMOT. We train our model on the train set of each dataset and test it on the validation set. The generalization ability of TCMP is shown in Table \ref{tab:ablation-generalization}. When trained and tested on the same dataset, which means that the data come from the same distribution, TCMP undoubtedly achieves competitive performance, demonstrating its effectiveness in capturing motion patterns native to the training data. TCMP achieves a HOTA score of $58.8\%$ on DanceTrack and $83.4\%$ on SportsMOT when trained on their respective datasets.

On cross-dataset evaluations, our proposed TCMP maintains robust performance overall. The models trained on SportsMOT and tested on DanceTrack and vice versa have a minimal drop in HOTA, demonstrating TCMP's capability to work with unseen data and, thus, generalize well.

\section{Conclusion}
\label{sec:conclusion}

Multi-object tracking (MOT) aims to identify and track individual objects across video frames. A crucial component of any MOT system is Motion prediction: forecasting future object positions based on past movement. While linear motion (constant speed and direction) is readily predicted using methods like the Kalman Filter, non-linear motion—characterized by sudden stops, accelerations, and turns—presents a significant challenge. Accurately predicting such non-linearities is vital for robust tracking in dynamic environments, but remains a difficult, open problem. Addressing this non-linear motion challenge is essential to advance MOT and enable its reliable use in real-world applications.

This paper addressed the critical challenge of non-linear motion prediction in multi-object tracking (MOT). This problem significantly limits the performance of many existing tracking systems in real-world scenarios.  We introduced the Temporal Convolutional Motion Predictor (TCMP), a novel framework leveraging a modified Temporal Convolutional Network (TCN) to efficiently and accurately model complex object trajectories. Unlike traditional methods that rely on linear motion assumptions (like the Kalman Filter) or computationally expensive autoregressive models, TCMP directly addresses the causality of motion prediction in online MOT, focusing on one-step-ahead forecasting to avoid cumulative errors. By generating these more accurate motion priors, TCMP is designed to significantly enhance the performance of the data association modules used in complete MOT frameworks.

Through comprehensive experiments on challenging benchmarks, including DanceTrack and SportsMOT, which feature highly non-linear and unpredictable object motion, we demonstrated that TCMP achieves state-of-the-art performance.  Specifically, TCMP significantly improves upon previous methods in key metrics such as HOTA, IDF1, and AssA, demonstrating superior tracking accuracy, identity preservation, and association accuracy.  Crucially, these improvements are achieved with a remarkably lightweight model – TCMP has a fraction of the parameters and computational cost of the prior state-of-the-art approach. This combination of high accuracy and efficiency underscores TCMP's capability for practical deployment in resource-constrained applications such as autonomous driving, surveillance, and robotics, where real-time performance is essential. Furthermore, comprehensive experiments on diverse benchmarks (DanceTrack, SportsMOT) demonstrate our model's superior performance in handling challenging scenarios.

\section*{Data Availability}

The datasets analyzed during the current study are publicly available in the following repositories and can be accessed via the provided persistent identifiers (PIDs):

\begin{itemize}
    \item \textbf{DanceTrack Dataset}. The DanceTrack benchmark dataset is publicly available in the official Github repository and can be accessed at \href{https://github.com/DanceTrack/DanceTrack}{https://github.com/DanceTrack/DanceTrack}. This dataset was used to generate results in Section \ref{sec:benchmark-evaluation}, Section \ref{sec:qualitative-dancetrack}, Table \ref{tab:exp-results-dancetrack}, Table \ref{tab:ablation-alpha}, Table \ref{tab:ablation-different-context-length}, Table \ref{tab:ablation-different-motion-model}, Table \ref{tab:exp-params-and-flops}, and Table \ref{tab:ablation-generalization}.

    \item \textbf{SportsMOT Dataset}. The SportsMOT benchmark dataset is publicly available in the official Github repository and can be accessed at \href{https://github.com/MCG-NJU/SportsMOT}{https://github.com/MCG-NJU/SportsMOT}. This dataset was used to generate results in Section \ref{sec:benchmark-evaluation}, Section \ref{sec:qualitative-sportsmot}, Table \ref{tab:exp-results-sportsmot}, Table \ref{tab:exp-params-and-flops}, and Table \ref{tab:ablation-generalization}.
\end{itemize}

\section*{Code Availability}
The source code used to generate the results presented in this study is publicly available on GitHub at \href{https://github.com/tcmp-tcn/tcmp}{https://github.com/tcmp-tcn/tcmp}.

\newpage
\begin{appendices}

\section{Pseudo-code of TCMP}
\label{sec:pseudo-code}

The pseudo-code of TCMP is shown in Algorithm \ref{alg:cosmo}. The input is a video sequence $V$, and an object detector $Det$, and a detection score thresholds $\tau_\text{high}$ and $\tau_\text{low}$. The output is the tracks $\mathcal{T}$ of the video, with each track $\mathcal{T}_i$ containing the bounding box $\mathbf{B}_i = (x, y, w, h)$ and identity $i$ of the object in each frame. 

For each frame in the video, we predict the detection boxes and scores using the detector $Det$. The detection boxes are separated into two $\mathcal{D}_\text{high}$ and $\mathcal{D}_\text{low}$ based on the detection score thresholds $\tau_\text{high}$ and $\tau_\text{low}$ we set earlier. For the detection boxes with scores higher than $\tau$, we put them into the high score detection boxes $\mathcal{D}_\text{high}$. and the detection boxes within $\tau_\text{high}$ and $\tau_\text{low}$ is put into $\mathcal{D}_\text{low}$ (line 5 to 13 in Algorithm \ref{alg:cosmo}). 

After separating the low-score detection boxes and the high-score detection boxes, TCMP $\mathcal{M}$ will perform motion prediction to predict the current position of each track in $\mathcal{T}$ in the current frame (lines 14 to 16 in Algorithm \ref{alg:cosmo}). 

The first association is performed between the high score detection boxes $\mathcal{D}_\text{high}$ and all the tracks $\mathcal{T}$ using the fusion of IoU and the Re-ID feature distances between the detection boxes $\mathcal{D}_\text{high}$ and the predicted box of tracks $\mathcal{T}$. Then, the Hungarian Algorithm \cite{kuhn1955hungarian} is used to perform the association step. The unmatched detections are stored in $\mathcal{D}_\text{remain}$ and the unmatched tracks in $\mathcal{T}_\text{remain}$ (line 20 and 21 in Algorithm \ref{alg:cosmo}). 

The second association is performed between the low score detection boxes $\mathcal{D}_\text{low}$ and the remaining tracks $\mathcal{T}_\text{remain}$ in the same fashion in the first association. The unmatched tracks in $\mathcal{T}_\text{re-remain}$ (lines 23 to 25 in Algorithm \ref{alg:cosmo}). Using IoU alone in the second association is important because the low-score detection boxes usually contain severe occlusion or motion blur, and appearance features are unreliable. 

Finally, The unmatched tracks will be deleted from the tracks $\mathcal{T}$.

\begin{algorithm}[ht]
\caption{Pseudo-code of TCMP}
\label{alg:cosmo}

\SetAlgoLined

\KwIn{A video sequence $V$; 
object detector $Det$; 
detection score thresholds $\tau_{\text{high}}, \tau_{\text{low}}$;
TCN Motion Predictor $\mathcal{M}$;}
\KwOut{Tracks $\mathcal{T}$ of the video.}

\textbf{Initialization:} $\mathcal{T} \gets \emptyset$\;

\For{frame $f$ in $V$}{
    \tcp{Detection}
    $\mathcal{D}_f \gets \mathcal{D}(f)$ \;
    $\mathcal{D}_{\text{high}} \gets \emptyset$; $\mathcal{D}_{\text{low}} \gets \emptyset$ \;
    
    \For{$d$ in $\mathcal{D}_f$}{
        \eIf{$d.\text{conf} > \tau_{\text{high}}$}{
             $\mathcal{D}_{\text{high}} \gets \mathcal{D}_{\text{high}} \cup \{d\}$\;
        }{
            \If{$\tau_{\text{low}} < d.\text{conf} < \tau_{\text{high}}$}{
                $\mathcal{D}_{\text{low}} \gets \mathcal{D}_{\text{low}} \cup \{d\}$\;
            }
        }
    }
    \;
    
    \tcp{Motion Prediction}
    \For{$t$ in $\mathcal{T}$}{
        $t \gets \mathcal{M}(t)$ \;
    }
    \;
    
    \tcp{First Association}
    Associate $\mathcal{T}$ and $\mathcal{D}_{\text{high}}$ using ReID and IoU\;
    $\mathcal{T}_{\text{remain}} \gets$ remaining tracks from $\mathcal{T}$\;
    $\mathcal{D}_{\text{remain}} \gets$ remaining detections from $\mathcal{D}_{\text{high}}$\;
    \;
    
    \tcp{Second Association}
    Associate $\mathcal{T}_{\text{remain}}$ and $\mathcal{D}    _{\text{low}}$ using IoU\;
    $\mathcal{T}_{\text{re-remain}} \gets$ remaining tracks from $\mathcal{T}_{\text{remain}}$\;

    $\mathcal{T} \gets \mathcal{T} \setminus T^{\text{re-remain}}$\;

    \For{$det$ in $\mathcal{D}_f^{\text{remain}}$}{
            $\mathcal{T} \gets \mathcal{T} \cup \{d\}$\;
    }
}
\Return{$\mathcal{T}$}\;
\end{algorithm}

\section{Qualitative Comparison}

\begin{figure*}[t!]
    \centering
    \begin{subfigure}[b]{0.48\textwidth}
        \centering
        \includegraphics[width=\textwidth]{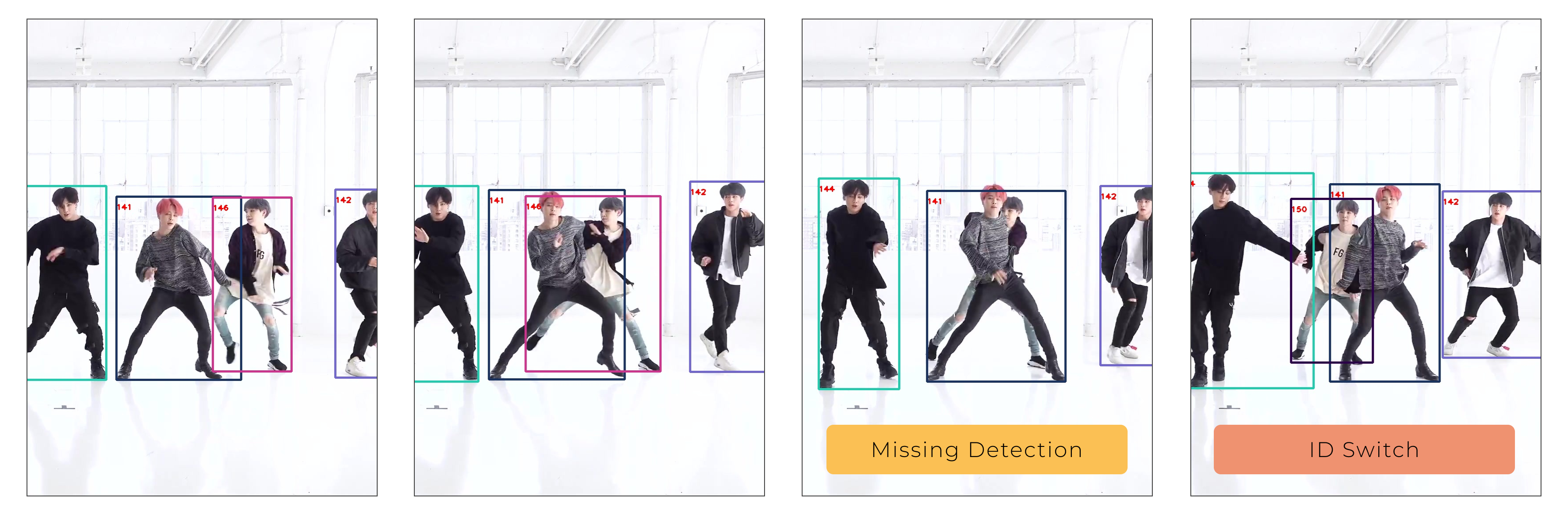}
        \caption{Case 1 (D$^2$MP)}
        \label{fig:qualitative-dancetrack-a}
    \end{subfigure} 
    \hfill
    \begin{subfigure}[b]{0.48\textwidth}
        \centering
        \includegraphics[width=\textwidth]{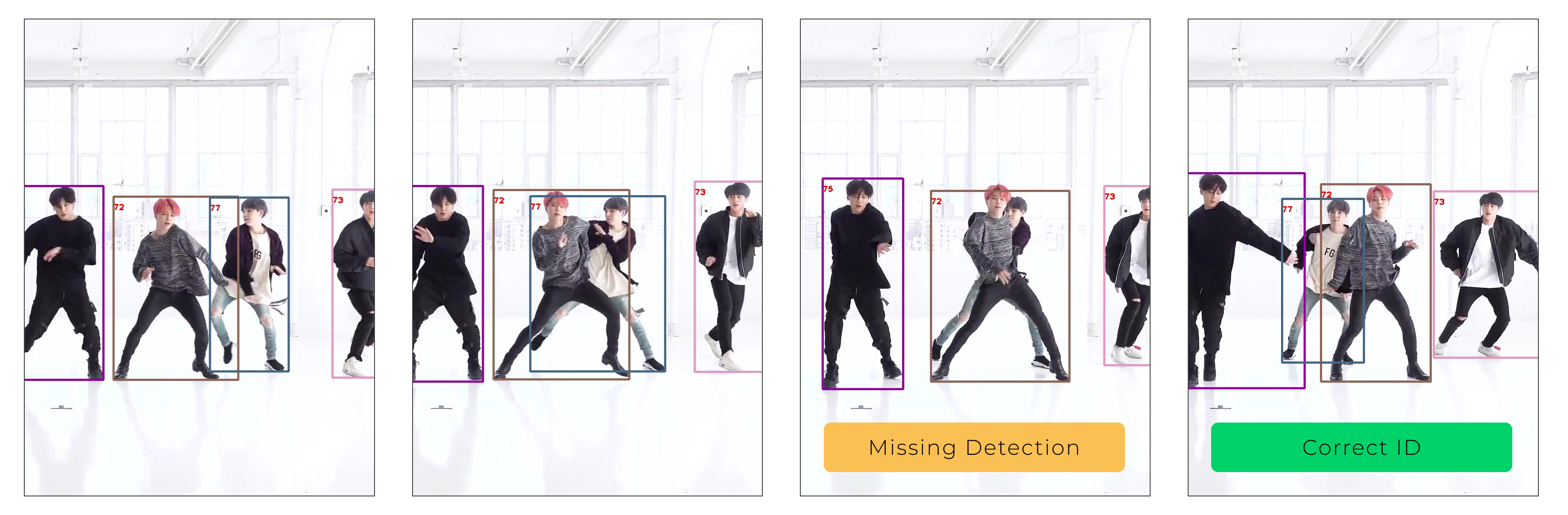}
        \caption{Case 1 (Ours)}
        \label{fig:qualitative-dancetrack-b}
    \end{subfigure}
    \hfill
    \begin{subfigure}[b]{0.48\textwidth}
        \centering
        \includegraphics[width=\textwidth]{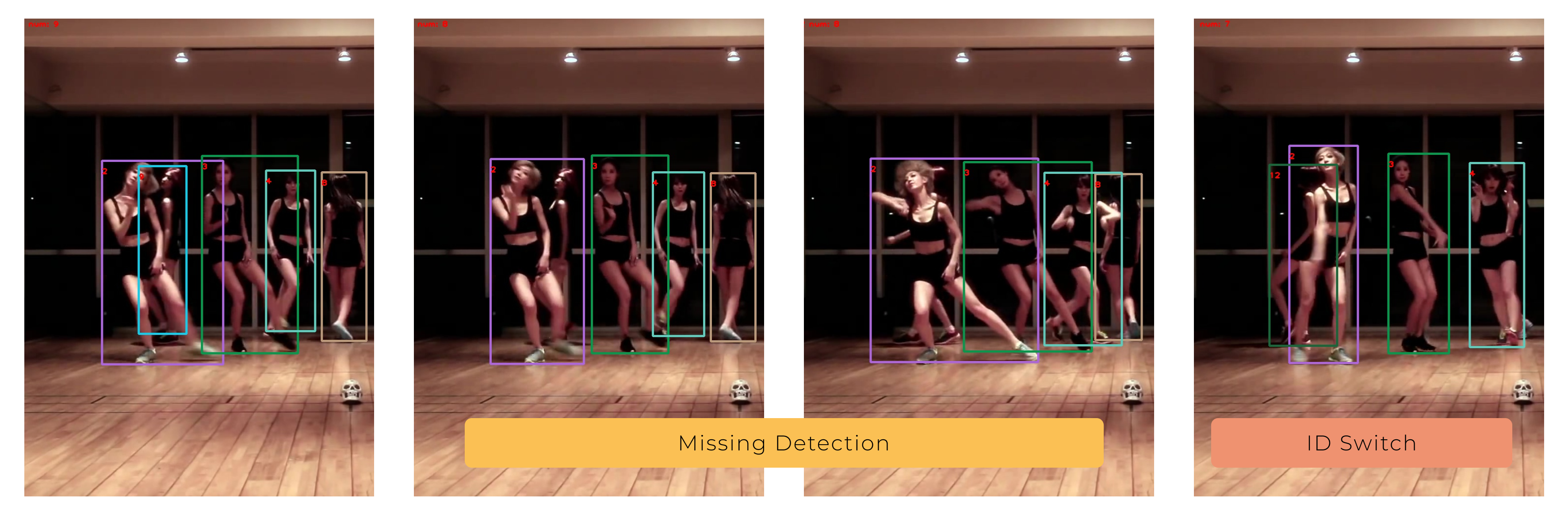}
        \caption{Case 2 (D$^2$MP)}
        \label{fig:qualitative-dancetrack-g}
    \end{subfigure} 
    \hfill
    \begin{subfigure}[b]{0.48\textwidth}
        \centering
        \includegraphics[width=\textwidth]{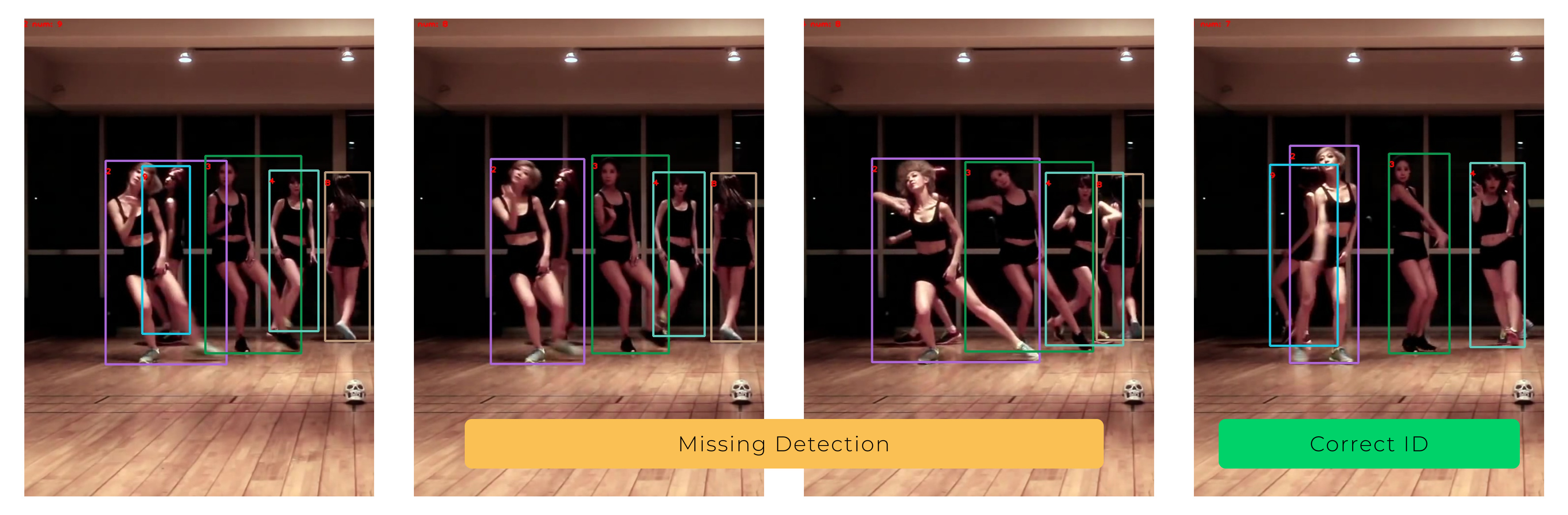}
        \caption{Case 2 (Ours)}
        \label{fig:qualitative-dancetrack-h}
    \end{subfigure}
    \hfill
    \begin{subfigure}[b]{0.48\textwidth}
        \centering
        \includegraphics[width=\textwidth]{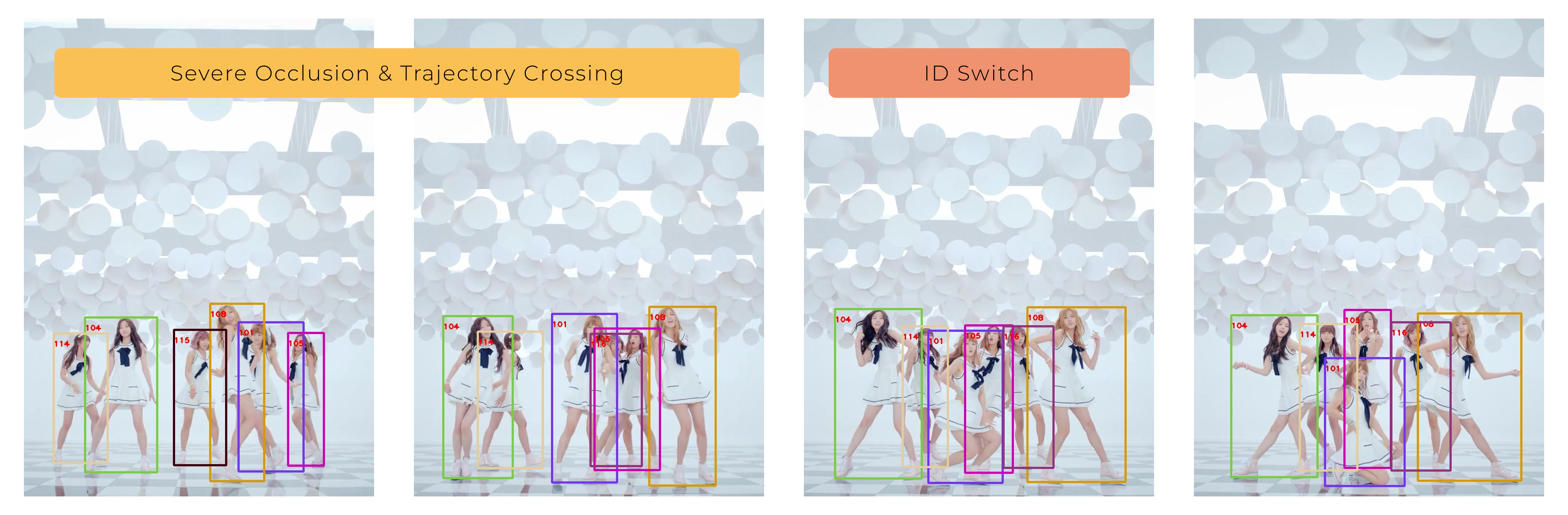}
        \caption{Case 3 (D$^2$MP)}
        \label{fig:qualitative-dancetrack-c}
    \end{subfigure}
    \hfill
    \begin{subfigure}[b]{0.48\textwidth}
        \centering
        \includegraphics[width=\textwidth]{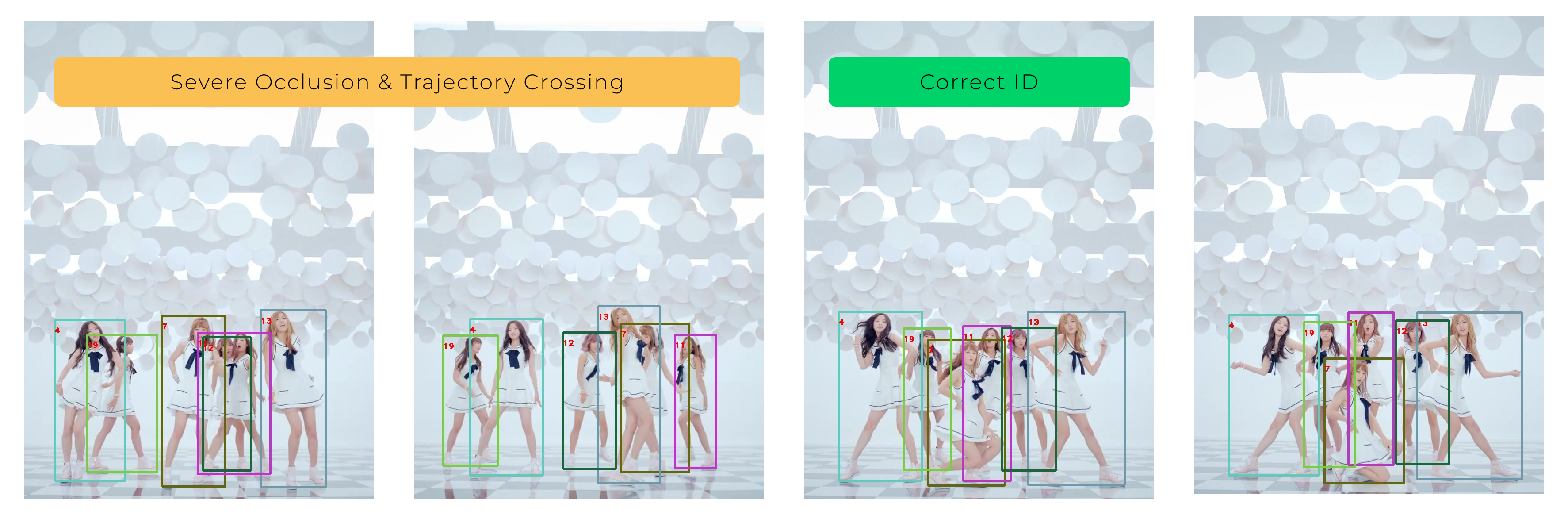}
        \caption{Case 3 (Ours)}
        \label{fig:qualitative-dancetrack-d}
    \end{subfigure}
    \hfill
    \begin{subfigure}[b]{0.48\textwidth}
        \centering
        \includegraphics[width=\textwidth]{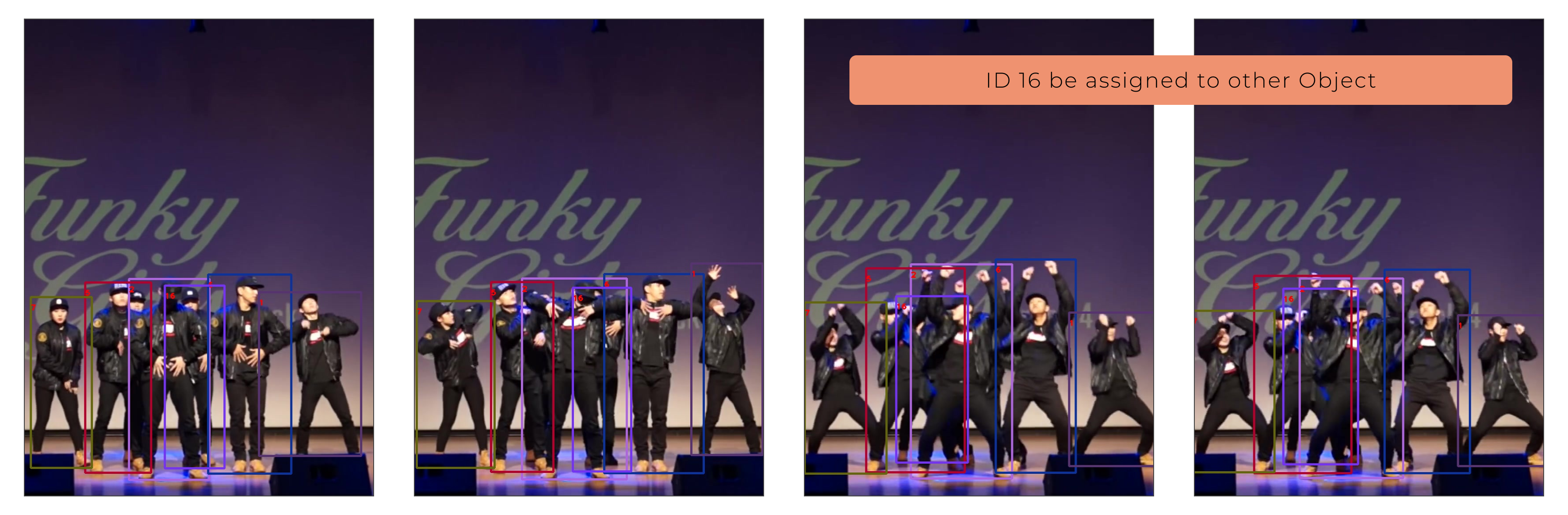}
        \caption{Case 4 (D$^2$MP)}
        \label{fig:qualitative-dancetrack-e}
    \end{subfigure}
    \hfill
    \begin{subfigure}[b]{0.48\textwidth}
        \centering
        \includegraphics[width=\textwidth]{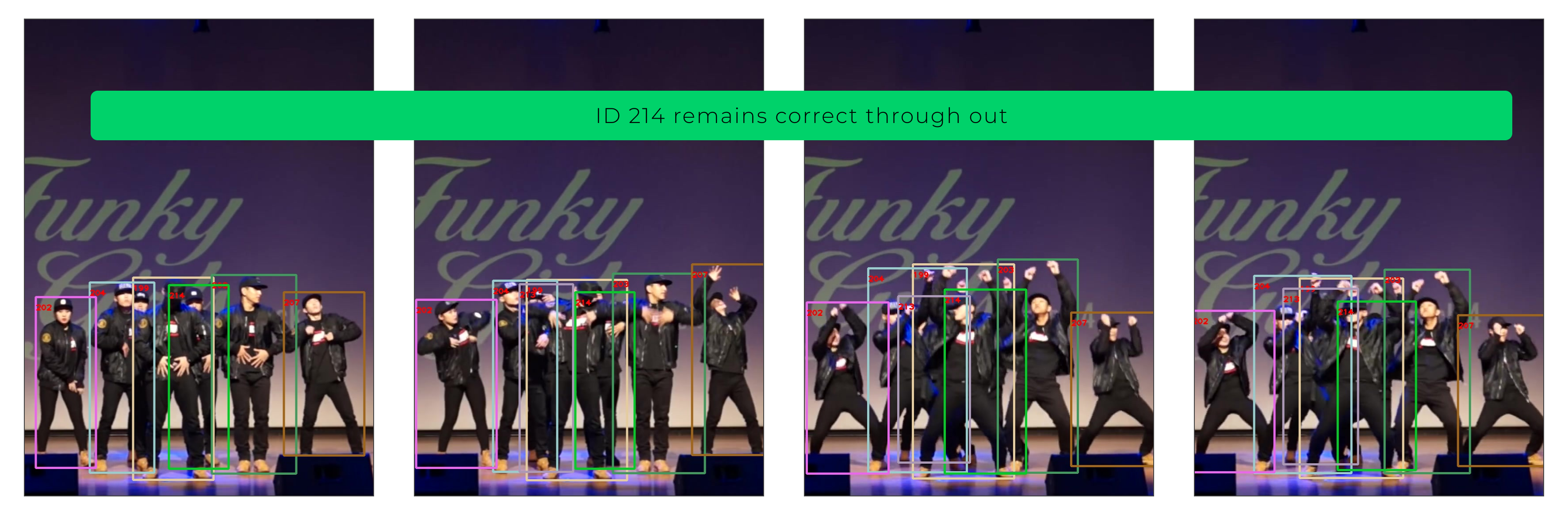}
        \caption{Case 4 (Ours)}
        \label{fig:qualitative-dancetrack-f}
    \end{subfigure}
    
    \caption{Qualitative comparison between using D$^2$MP and TCMP as the motion model on the DanceTrack test set. (a), (c), (e) and (g) represent the results using D$^2$MP as the motion model. (b), (d), (f) and (h) represent the results using TCMP as the motion model. Each pair of rows shows the comparison of the results for one sequence. Boxes of the same color and number represent the same ID. Best viewed in color and zoom-in}
        \label{fig:qualitative-dancetrack}
\end{figure*}

\begin{figure*}[t!]
    \centering
    \begin{subfigure}[b]{0.48\textwidth}
        \centering
        \includegraphics[width=\textwidth]{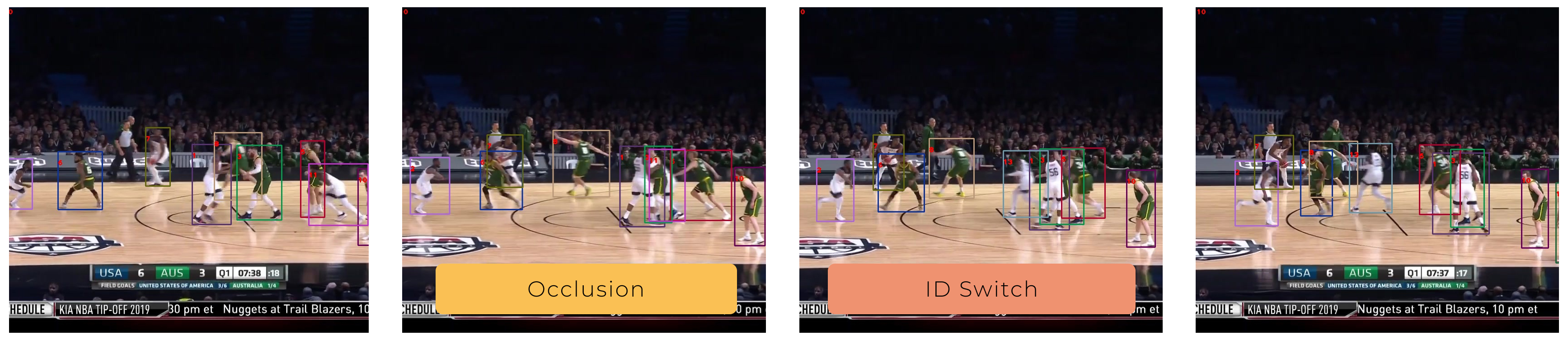}
        \caption{Basketball (D$^2$MP)}
        \label{fig:qualitative-sportsmot-a}
    \end{subfigure} 
    \hfill
    \begin{subfigure}[b]{0.48\textwidth}
        \centering
        \includegraphics[width=\textwidth]{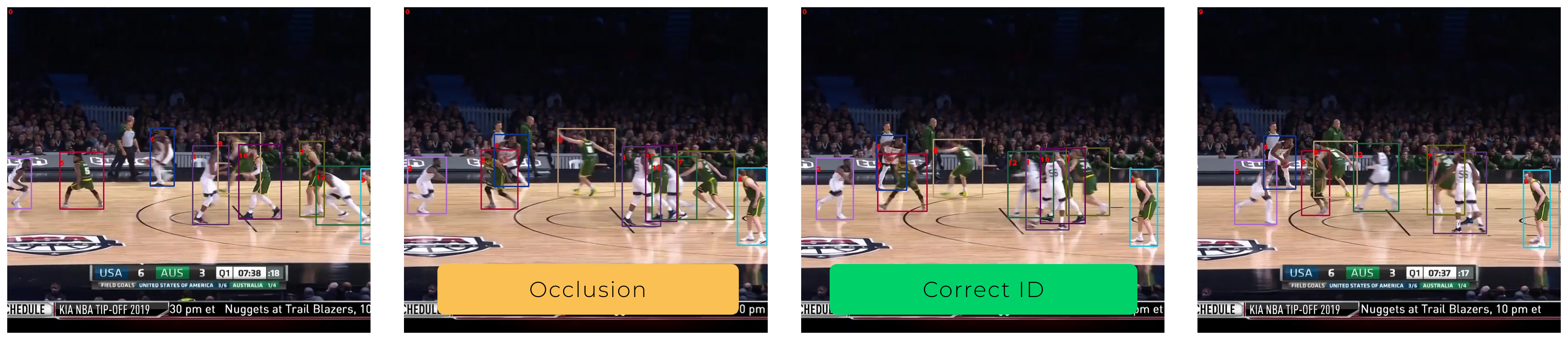}
        \caption{Basketball (Ours)}
        \label{fig:qualitative-sportsmot-b}
    \end{subfigure}
    \hfill
    \begin{subfigure}[b]{0.48\textwidth}
        \centering
        \includegraphics[width=\textwidth]{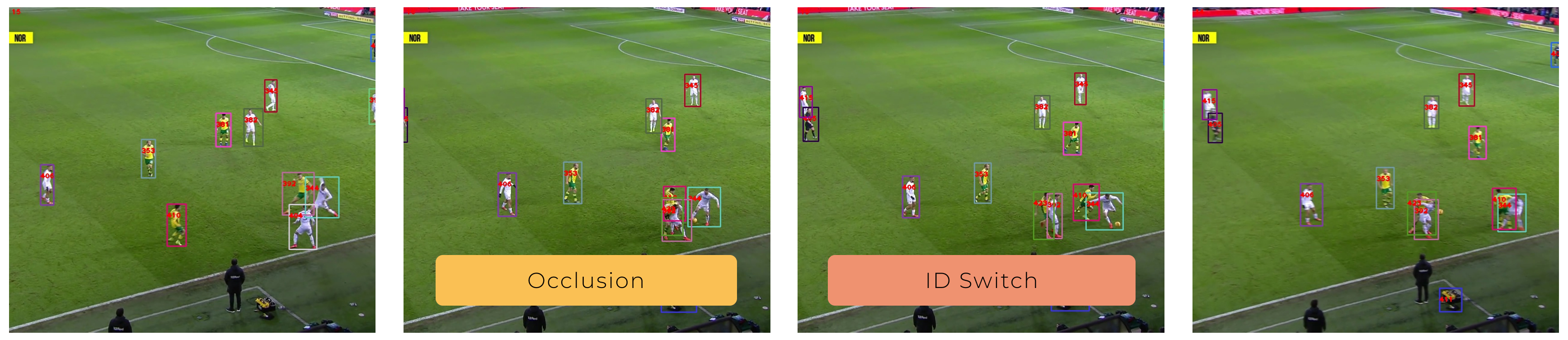}
        \caption{Football (D$^2$MP)}
        \label{fig:qualitative-sportsmot-c}
    \end{subfigure} 
    \hfill
    \begin{subfigure}[b]{0.48\textwidth}
        \centering
        \includegraphics[width=\textwidth]{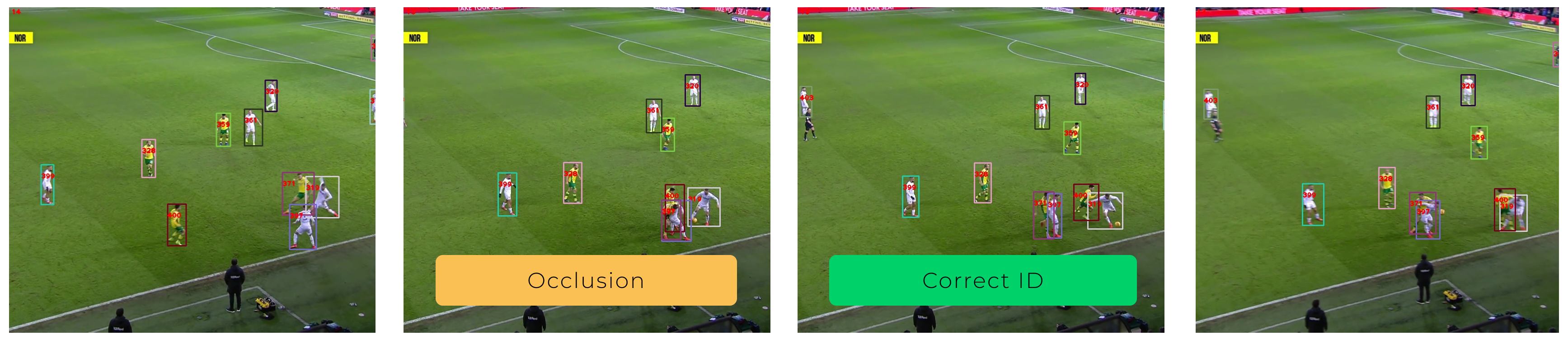}
        \caption{Football (Ours)}
        \label{fig:qualitative-sportsmot-d}
    \end{subfigure}
    \hfill
    \begin{subfigure}[b]{0.48\textwidth}
        \centering
        \includegraphics[width=\textwidth]{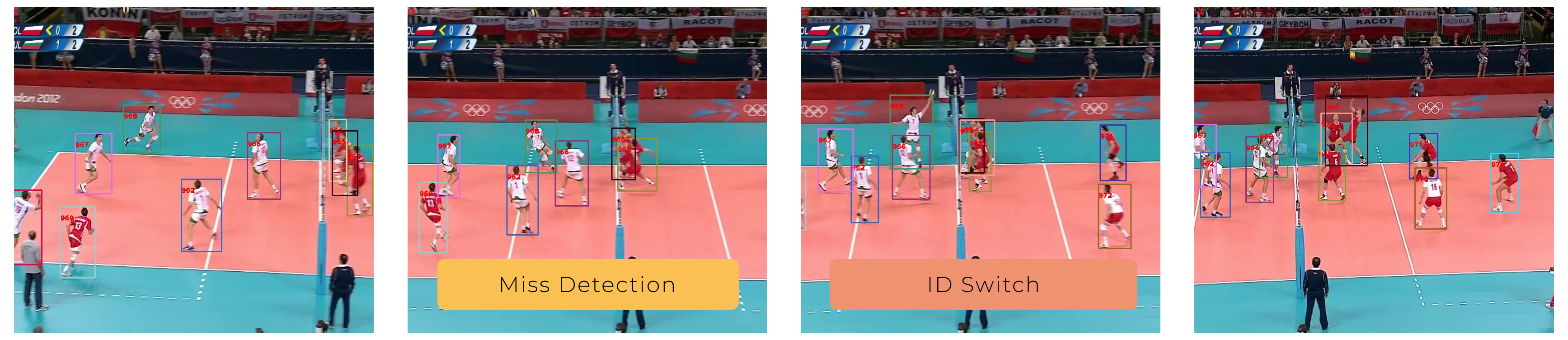}
        \caption{Volleyball (D$^2$MP)}
        \label{fig:qualitative-sportsmot-e}
    \end{subfigure}
    \hfill
    \begin{subfigure}[b]{0.48\textwidth}
        \centering
        \includegraphics[width=\textwidth]{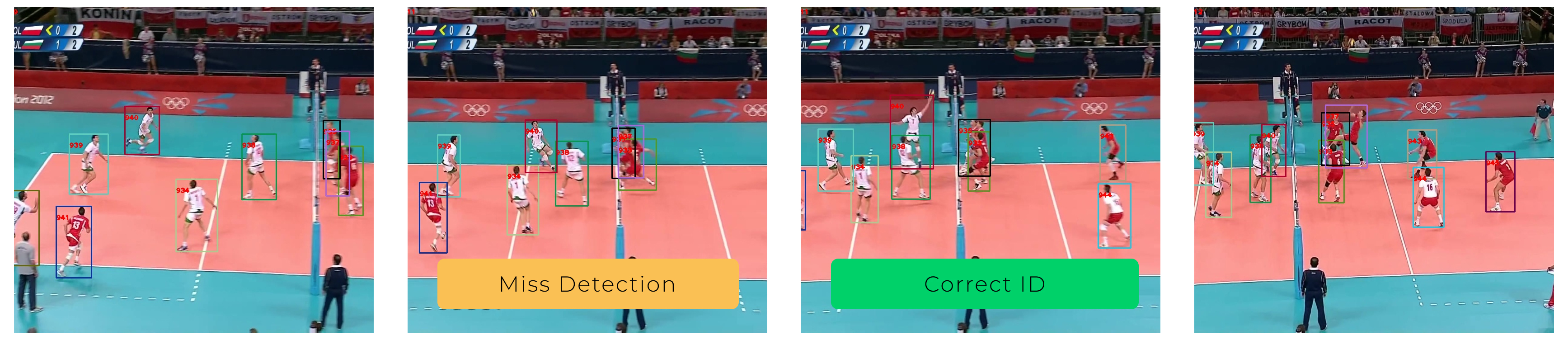}
        \caption{Volleyball (Ours)}
        \label{fig:qualitative-sportsmot-f}
    \end{subfigure}
    
    \caption{Qualitative comparison between using D$^2$MP and TCMP as the motion model on the SportsMOT test set. (a), (c), and (e) represent the results using D$^2$MP as the motion model. (b), (d), and (f) represent the results using TCMP as the motion model. Each pair of rows shows the comparison of the results for one sequence. Boxes of the same color and number represent the same ID. Best viewed in color and zoom-in}
        \label{fig:qualitative-sportsmot}
\end{figure*}

\subsection{Qualitative comparison on DanceTrack}
\label{sec:qualitative-dancetrack}

Figure \ref{fig:qualitative-dancetrack} shows some samples on the test set of DanceTrack. These 4 cases present 4 different difficult scenarios in Multi-object Tracking. 

\textbf{Case 1: Short-Term Missed Detections}. Case 1 shows a situation where the detector fails to detect an object, leading to some detections being missed. The motion model must be capable of continuing to predict the object's movement so that when the object reappears in the next frame, it can be matched to its previous ID. In the sequence shown in Figure \ref{fig:qualitative-dancetrack-a}, DiffMOT assigns a different ID to the object after a miss-detection (ID146 $\rightarrow$ ID150). Our model, however, can still correctly predict and assign the correct ID to the previously miss-detected object (Figure \ref{fig:qualitative-dancetrack-b}).

\textbf{Case 2: Long-Term Missed Detections}. Case 2 shows a situation similar to Case 1, but the object is missed for a longer duration, during a more severe detection failure. Such scenarios amplify the difficulty in maintaining identity consistency, testing the ability of "predicting in the dark", where there is no new observation for feedback. In the sequence shown in Figure \ref{fig:qualitative-dancetrack-g}, DiffMOT assigns a different ID to the object (ID9 $\rightarrow$ ID12). In contrast, our model still assigns the correct ID to the object (Figure \ref{fig:qualitative-dancetrack-h}).

\textbf{Case 3: Severe occlusion with crossing trajectories}. Case 3 shows a situation where there is severe occlusion and objects' trajectories cross each other. Motion models need to "remember" the direction of each object and predict more precise bounding boxes so that they can be associated correctly in the association step. In Figure \ref{fig:qualitative-dancetrack-c}, DiffMOT assigns a different ID to the object (ID146 $\rightarrow$ ID150), while in Figure \ref{fig:qualitative-dancetrack-d}, TCMP remains correct ID for all objects.

\textbf{Case 4: Close overlap with complex motions}. Case 4 shows objects are highly close/overlapping and have complex motion. In Figure \ref{fig:qualitative-dancetrack-e}, DiffMOT assigns an ID of one object to another (ID16 tracks the object on the right side of ID2 and switches to another object that is on the left of ID2). In contrast, in Figure \ref{fig:qualitative-dancetrack-f}, TCMP keeps ID217 (the same object as ID16 in Figure \ref{fig:qualitative-dancetrack-e}) on the same object.

\subsection{Qualitative comparison on SportsMOT}
\label{sec:qualitative-sportsmot}

Figure \ref{fig:qualitative-sportsmot} shows some samples on the test set of the SportsMOT dataset. Each row represents different sports in the SportMOT datasets, namely Basketball (Figure \ref{fig:qualitative-sportsmot-a} and Figure \ref{fig:qualitative-sportsmot-b}), Football (Figure \ref{fig:qualitative-sportsmot-c} and Figure \ref{fig:qualitative-sportsmot-d}), and Volleyball (Figure \ref{fig:qualitative-sportsmot-e} and Figure \ref{fig:qualitative-sportsmot-f}).

\end{appendices}

\bibliography{bibliography}% common bib file

\end{document}